\pdfoutput=1

\documentclass[11pt]{article}

\usepackage[]{acl}

\usepackage{times}
\usepackage{latexsym}

\usepackage{hyperref}
\usepackage{url}
\usepackage{multirow}
\usepackage{graphicx}
\usepackage{comment}
\usepackage{amssymb}
\usepackage{amsmath}
\usepackage{url}

\usepackage[T1]{fontenc}

\usepackage[utf8]{inputenc}

\usepackage{microtype}

\title{Semi-Supervised Dialogue Abstractive Summarization via High-Quality 
Pseudolabel Selection }

 \author{
Jianfeng He{$^{1,2}$}\thanks{~The work was done during an AWS AI Labs internship.}~, Hang Su{$^1$}, Jason Cai{$^1$}, Igor Shalyminov{$^1$}, Hwanjun Song{$^1$}, Saab Mansour{$^1$}
\\ 
   {$^1$} AWS AI Labs\\
 {$^2$} Virginia Tech\\
 jianfenghe@vt.edu, \{shawnsu, cjinglun, shalymin, saabm\}@amazon.com
 }

\begin{document}
\maketitle
\begin{abstract}
Semi-supervised dialogue summarization (SSDS) leverages model-generated summaries to reduce reliance on human-labeled data and improve the performance of summarization models. While addressing label noise, previous works on semi-supervised learning primarily focus on natural language understanding tasks, assuming each sample has a unique label. However, these methods are not directly applicable to SSDS, as it is a generative task, and each dialogue can be summarized in different ways. 
In this work, we propose a novel scoring approach, SiCF, which encapsulates three primary dimensions of summarization model quality: Semantic invariance (indicative of model confidence), Coverage (factual recall), and Faithfulness (factual precision). Using the SiCF score, we select unlabeled dialogues with high-quality generated summaries to train summarization models. Comprehensive experiments on three public datasets demonstrate the effectiveness of SiCF scores in uncertainty estimation and semi-supervised learning for dialogue summarization tasks. Our code is available at \url{https://github.com/amazon-science/summarization-sicf-score}.
\end{abstract}

\section{Introduction}
Dialogue summarization generates concise summaries of dialogues, helping users quickly understand key points without navigating through complex contexts~\cite{feng2021survey}. This study prioritizes abstractive summarization, which offers more flexibility than extractive approaches~\cite{gupta2019abstractive, wong2008extractive}. Despite its wide applicability in scenarios like meetings and casual conversations, dialogue summarization faces challenges such as scarcity of annotations and high annotation costs. However, the proliferation of pre-trained models and unlabeled dialogues offers a solution. In this paper, we explore \textbf{S}emi-\textbf{S}upervised \textbf{D}ialogue \textbf{S}ummarization (SSDS)~\cite{chen2021simple}, aiming to enhance dialogue summarization by a small labeled dataset alongside a large collection of unlabeled dialogues.

Previous SSDS research~\cite{chen2021simple} has used data augmentation to increase the size of both labeled and unlabeled dialogue datasets, but the issue of pseudolabel noise has been largely overlooked. Specifically, an initial model fine-tuned on labeled samples is used to generate pseudolabels for unlabeled samples. Then, the unlabeled samples and their pseudolabels are used to train the semi-supervised model~\cite{rizve2021defense}. However, the imperfections of the initial model can lead to pseudolabel noise, such as hallucination and missing key information. Pseudolabel noise is a big concern in semi-supervised learning because training on pseudolabels with significant noise can deteriorate model performance. Thus, we aim to address pseudolabel noise in SSDS in this research.

Many existing solutions for pseudolabel noise estimation and mitigation
are designed for understanding tasks (e.g., classification~\cite{cordeiro2020survey}), and they are not directly applicable to SSDS due to inherent diversity of ground truth summaries. Specifically, these solutions, such as Mix-Up in Mix-Match~\cite{berthelot2019mixmatch}, assume each sample has a unique label, representing a single attribute like a semantic class. 
In contrast, SSDS is a generation task where each dialogue can be summarized in different ways. For instance, summaries like ``the audience is happy to hear the news'' and ``the news makes the audience glad'' convey the same message with different wording. As a result, SSDS, like other generation tasks, does not have a unique label per sample. This distinction makes previous pseudolabel noise solutions unsuitable for SSDS. Thus, we need a new and generalized pseudolabel noise measurement solution that considers label diversity in SSDS, without relying on ground truth summaries, as unlabeled dialogues lack them.

\begin{figure*}[tbh]
\centering
\includegraphics[width=1.0\textwidth]{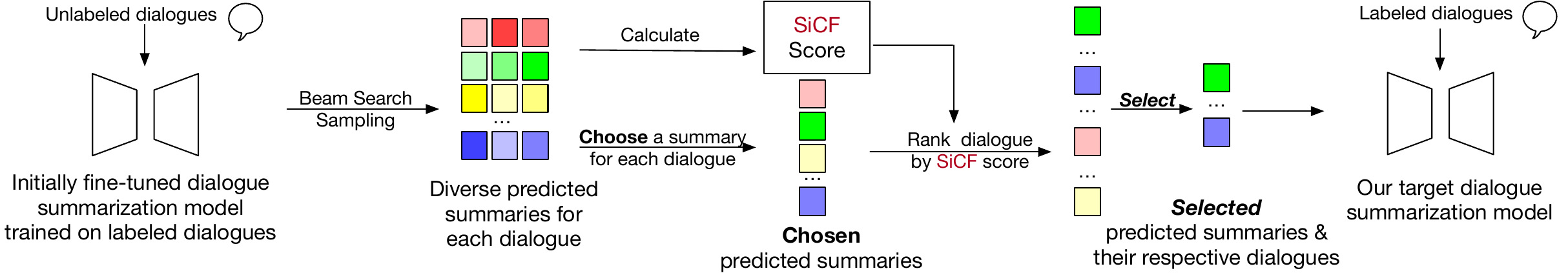}
\caption{
A global view of our SSDS framework using the semantic invariance, coverage, and faithfulness combined score (SiCF). Each row in the colored matrix represents diverse predicted summaries for a dialogue. For each unlabeled dialogue, the predicted summary closest to mean embedding is chosen. We then rank the chosen predicted summaries by the SiCF scores and select a portion of them. The selected <unlabeled dialogues, pseudolabels> and all human-labeled pairs are used for our target model learning.
The detailed SSDS framework is outlined in Sec.~\ref{sec:overview_ssds}. 
}
\label{fig:global_view}
\end{figure*}

To achieve this, we propose assessing pseudolabel quality: a predicted summary with higher quality indicates less noise in pseudolabels.  We thus propose the SiCF score, which measures summary quality based on common characteristics of high-quality summaries, such as model confidence, information coverage, and faithfulness to the original dialogues. SiCF comprises three components: ``\underline{S}emantic \underline{i}nvariance'' assesses model confidence at the text level, ``\underline{C}overage'' evaluates key information captured at the word level, and ``\underline{F}aithfulness'' measures alignment with the original dialogues at the sentence level. 
As shown in Figure~\ref{fig:global_view}, we then rank and select the unlabeled dialogues with high-quality pseudolabels as indicated by the SiCF score. 
Besides, since uncertainty estimation is a representative way to estimate the model prediction quality~\cite{gawlikowski2023survey,he2023uncertainty} and Bayesian Neural Network (BNN) is an effective uncertainty estimation method~\cite{mukhoti2023deep}, we propose a variant-length multi-label BNN for our SiCF score.
Our contributions are as below. 
\begin{itemize}    
    \item We propose the SiCF score framework to measure the quality of predicted summaries based on these three key characteristics. To the best of our knowledge, we are the first to comprehensively evaluate summary quality without relying on ground truth summaries.

    \item We introduce a variant-length multi-label BNN uncertainty estimation technique used in the SiCF score. In contrast, conventional BNN~\cite{mukhoti2023deep} is designed for fixed-length single-label cases, which do not align with the requirements of our task.
\end{itemize}

\section{Related Work}
\textbf{Semi-supervised text summarization.} 
Please refer to Sec.~\ref{app:rel_sum_qua} for the related work of Semi-supervised text summarization.

\noindent\textbf{Semi-supervised dialogue summarization.} Semi-supervised dialogue summarization is also under-explored, although some works focus on guiding dialogue summarization \cite{liu2021controllable}, improving model performance via human feedback \cite{chen2022human}, and enhancing factual consistency between ground-truth and generated summaries \cite{chen2021dialogue}.

In terms of semi-supervised extractive dialogue summarization, \citet{mishra2023llm} employ GPT 3.5 for quality assessment based on token probabilities. \citet{zhuang2023self} introduce self-supervised pre-training to enhance BERT's ability to contextualize dialogue representations.

Regarding our focus, that is the semi-supervised abstractive dialogue summarization, CODA \cite{chen2021simple} is proposed to address SSDS using data augmentation. While data augmentation can expand the size of both labeled and unlabeled data, it overlooks challenges posed by pseudolabel noise, a prevalent issue in semi-supervised learning. Unlike previous SSDS models that overlooked pseudolabel noise, our goal is to enhance SSDS performance by measuring pseudolabel quality and effectively eliminating unreliable pseudolabels.

\noindent\textbf{Solution of pseudolabel noise in semi-supervised learning.} Many methods have been proposed for label noise in natural language understanding tasks ~\cite{cordeiro2020survey,berthelot2019mixmatch,he2023zero,lei2022uncertainty}. However, most of these methods are not directly applicable to SSDS, because this generation task has diverse ground truth summaries for each dialogue. 
While some of these methods have potential to be applied towards SSDS, like teacher-student knowledge distillation model for noisy text summarization ~\cite{liu2020noisy},  they do not consider diversity of summaries within SSDS.~\citet{rizve2021defense} have a similar task setting to ours, but their task is multi-label image classification, which still provides a unique label for each image. 
~\citet{wan2023better} focus solely on model generation without considering the interaction with context (e.g., dialogue in our task). In contrast, we consider both model prediction itself by semantic invariance and the relation between generations and context via coverage and faithfulness. 

As for injecting noise into the dialogues or pseudolabels~\cite{he2019revisiting}, it focuses on improving the model's robustness through training with this injected noise and aims to mitigate the impact from noise. In contrast, our work aims to measuring the extent of sample noise. Furthermore, they need to retrain the model, and the added noise might degrade the model's performance. In contrast, our work does not require retraining the model and will not harm its performance.

\noindent\textbf{Summary quality.} Please refer to Sec.~\ref{app:rel_sum_qua} for the related work of ``Summary quality''.

\section{Problem Setting}
We are given a dialogue set $D^{l}$ with annotated summaries and a dialogue set $D^{u}$ without annotations.  $D^{l}$ and $D^{u}$ belong to the same domain, as our focus is not domain generalization. Given a pretrained dialogue summarization model $G_{0}$, we leverage $D^{l}$ and some or all data of $D^{u}$ to fine-tune $G_{0}$, obtaining a target dialogue summarization model $\hat{G}$. We aim to accurately evaluate the quality of pseudolabels for unlabeled samples, enabling us to select higher-quality unlabeled data for training an improved model $\hat{G}$. 

\section{Our Model}
\begin{figure*}[tbh]
\centering
\includegraphics[width=0.95\textwidth]{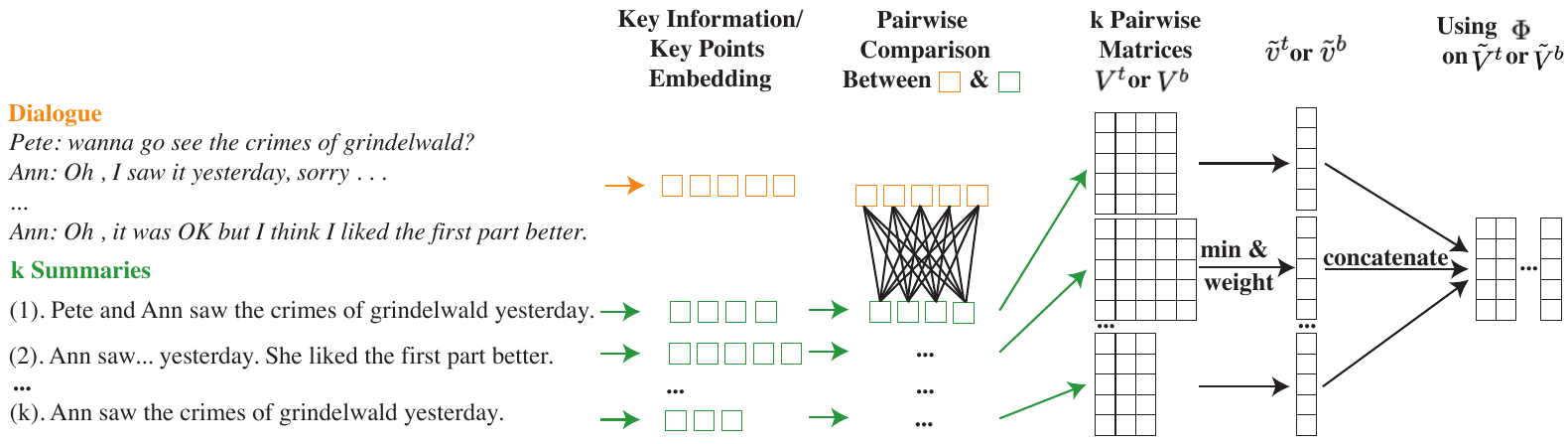}
\caption{
The global view of our coverage and faithfulness scores in our SiCF score.
}
\label{fig:cov_fai}
\end{figure*}

\subsection{Overview of SSDS via our SiCF Score}
\label{sec:overview_ssds}
The proposed framework for SSDS with SiCF scores is shown in Figure~\ref{fig:global_view}. 
We begin with a pretrained dialogue summarization model $G_{0}$, such as DialogLED~\cite{zhong2022dialoglm}~\footnote{We select DialogLED for its superior performance and hardware efficiency, facilitating accessibility for other researchers using our setup.}. We first fine-tune $G_{0}$ on dialogue-summary pairs in $D^{l}$. 
To assess pseudolabel quality, we use uncertainty estimation, which is an effective way to measure the model prediction quality~\cite{gawlikowski2023survey}. 
Bayesian Neural Network (BNN) is an effective method for uncertainty estimation and is often approximated by ensemble~\cite{gal2016dropout}. We generate $k$ diverse summaries for each unlabeled dialogue from $D^{u}$ by beam search sampling~\cite{vijayakumar2016diverse}.  Next, based on these diverse summaries, we calculate our SiCF score for each unlabeled dialogue to evaluate its summary quality. This score includes three aspects: Semantic invariance, Coverage, and Faithfulness. Moreover, we \textbf{choose} a summary for each dialogue based on the embedding that is closest to the mean of its all diverse summary embeddings. Next, we rank and \textbf{select} high-quality dialogue-pseudolabel pairs based on the SiCF scores.
Finally, we fine-tune $G_{0}$ with the labeled dialogues and the \textit{selected} unlabeled dialogues with pseudolabels to train the target dialogue summarization model $\hat{G}$.

Our work focuses on (1) how to obtain SiCF scores that accurately measure the quality of predicted summaries based on uncertainty estimation, and (2) how to use SiCF scores to select high-quality unlabeled dialogues and then improve $\hat{G}$. 

We detail the reasons for choosing semantic invariance, coverage, and faithfulness in Sec.~\ref{sec:app_resons_three}.

\subsection{SiCF Score: Semantic Invariance}

\citet{kuhn2023semantic} propose semantic uncertainty based on semantic invariance for text generation quality evaluation. 
Higher semantic invariance means smaller semantic divergence between the $k$ diverse generations of a sample, and thus indicating a higher quality in generations' semantics.  
However, \citet{kuhn2023semantic} needs to cluster diverse generations for each sample, which is time-consuming for large sample size. 

Different from ~\citet{kuhn2023semantic}, we propose a variance-based method to measure the semantic invariance without clustering. 
This is because variance is also an effective uncertainty estimation method when the task has no unique ground truth (e.g., text summarization)~\cite{var2uncertainty}.
Specifically, given a dialogue from $D^{u}$ with $k$ diverse predicted summarizations $s=\{s_{1}, s_{2}, ..., s_{k}\}$, we use a  pretrained encoder model (e.g., RoBERTa~\cite{liu2019roberta}) to produce their text embeddings as $e_{1}, e_{2}, ..., e_{k}$, we get a semantic invariance score $\lambda_{SeIn}$ for the unlabeled dialogue by the variance of its diverse summary embeddings as follows,

\begin{equation}
\label{eq:semantic_inv}
\lambda_{SeIn} = var(e_{1}^{s}, e_{2}^{s}, ..., e_{k}^{s})
\end{equation}

Our variance operation is more efficient than~\citet{kuhn2023semantic}, as variance achieves a time complexity of $O(k)$ compared to their $O(k^2)$.

\subsection{SiCF Score: Coverage}
\label{sec:cov}
Since a good summary typically covers key details in a dialogue, we design a coverage score to measure the quality of a summary. Unlike coverage in~\citet{huang2023swing}, ours does not depend on ground truth summaries.

To get our coverage score, we need to extract key details from a dialogue. Based on our observation (illustrated in Sec.~\ref{sec:compare_pos_ner}) and conciseness of summaries, we use the nouns in a dialogue to represent its key details. This is because the nouns in a dialogue carry the information to distinguish themselves from other dialogues.

Concretely, we use a pretrained POS tagging model (i.e., Flair~\cite{akbik2019flair}) to extract nouns from a dialogue as its key details. Also, key details (nouns) can be extracted from a corresponding summary for comparison. 
We use the $T^{d}=[t_1^{d},t_2^{d},..., t_p^d]$ to represent a sequence of noun embedding from a dialogue $d$, which has $p$ nouns. 
Similarly, we use the $T^{s}=[t_1^{s},t_2^{s},..., t_q^s]$ to represent a sequence of noun embedding of a summary $s$. As shown in Figure~\ref{fig:cov_fai}, we then calculate the similarity matrix $V^{t}\in\mathbb{R}^{p \times q}$ 
between noun embeddings $T^{d}$ and $T^{s}$, 
\begin{equation}
\label{eq:noun_sim}
V^{t}= Dist(T^{d}, T^{s})
\end{equation}
where $Dist$ is pair-wise Euclidean distance. We choose Euclidean distance, as we expect a small value means high quality. 
A smaller value in  $V^{t}$ means better similarity between a noun of the dialogue and a noun of its one summary. Thus, we then apply row-level min operation on $V^{t}$ to get coverage vector $\hat{v}^t \in \mathbb{R}^{p}$, 
that is $\hat{v}^t = min(V^{t})$,
where each element indicates the coverage degree between a predicted summary and a noun in the dialogue. We further weight the $\hat{v}^t$ as $\tilde{v}^t=w^t\cdot\hat{v}^t$, where $\tilde{v}^t\in\mathbb{R}^{p}$. The $w^{t}\in\mathbb{R}^{p}$ is the weight of each noun in dialogue, measured by noun's occurrence. 
Since speaker names in dialogues are proper nouns that often repeat, we take the maximum occurrence of proper nouns as 1 to prevent bias in the model due to speaker names.

Since we have $k$ diverse generated summaries for each dialogue, we can have $\tilde{V}^{t}=[\tilde{v}_{1}^{t}, \tilde{v}_{2}^{t}, ..., \tilde{v}_{k}^{t}]\in \mathbb{R}^{k \times p}$, where $\tilde{v}_{i}^{t}$ is a coverage vector of $i$-th diverse generated summary for the dialogue. Since coverage score should be a scalar, we use a function $\Phi$ to get a coverage score $\lambda_{cov}$,
\begin{equation}
\label{eq:coverage}
\lambda_{cov}=\Phi(\tilde{V}^t)
\end{equation}
where $\Phi$ can be mean, BNN, or their combination (m+BNN), which will be introduced in Sec.~\ref{sec:vec2score}.

\subsection{SiCF Score: Faithfulnesss}
\label{sec:fai}

Because a good summary should adhere to the key point of the dialogue, we consider faithfulness, which is the adherence degree between the key points of a dialogue and its summaries.
However, using details (e,g., nouns) as key points may omit the connection of state words, like "not" and "disagrees". But using the text-level embedding is too general to miss the fine-grained information, such as SummaC~\cite{laban2022summac}. As a result, we consider sentence-level key points, because it keeps both state words and fine-grained information.  Unlike faithfulness in~\citet{huang2023swing}, ours does not depend on ground truth summaries.

Specifically, given a dialogue with $h$ sentences and a predicted summary with $z$ sentences, we have a sequence of dialogue sentence embeddings $B^{d}=[b_1^{d},b_2^{d},..., b_{h}^d]$ and a sequence of summary sentence embeddings $B^{s}=[b_1^{s},b_2^{s},..., b_{z}^{s}]$ for them by an encoder of a pretrained Natural Language Inference (NLI) model, which is effective in faithfulness-check models (e.g. FactCC~\cite{kryscinski2021evaluating} and SummaC~\cite{laban2022summac}). As shown in Figure~\ref{fig:cov_fai}, we utilize the pretrained NLI model to obtain a faithfulness matrix $V^b\in\mathbb{R}^{h\times z}$ as follows.
\begin{equation}
\label{eq:point_sim}
V^{b}=NLI(B^{d}, B^{s})
\end{equation}
The ${h\times z}$ shape is built by pair-wise comparing $h$ sentences in a dialogue to $z$ sentences in a summary.
Each element in $NLI(B^{d}, B^{s})$ is obtained by first calculating the NLI negative and positive results between $i$-th dialogue sentence embedding $b_i^{d}$ and $j$-th summary sentence embedding $b_j^{s}$, followed by returning the NLI result of these two sentences. The NLI result in our work is  negative score subtracting the positive score, which is similar to SummaC.  As a result, a smaller element in $V^b$ means better faithfulness between a dialogue sentence and a summary sentence. 

Next, similar to coverage score, we apply row-level min operation on $V^b$ and have $\hat{v}^b\in\mathbb{R}^{h}$. Each element in $\hat{V}$ indicates the faithfulness agreement between the summary sentences and a dialogue sentence. We further weight $\hat{v}^b$ as $\tilde{v}^b = \hat{v}^b \cdot w^b$, where $w^b \in \mathbb{R}^{h}$ has each element as the noun occurrences in a dialogue sentence. 
We also limit the proper noun occurrences to a maximum of 1 because names in the dialogue are frequently mentioned and less significant than other nouns.

Since there are $k$ summaries for each dialogue, we can then obtain $\tilde{V}^b=[\tilde{v}_1^b, \tilde{v}_2^b, ..., \tilde{v}_k^b]\in\mathbb{R}^{k\times h}$. 
Finally, similar to Eq.~\ref{eq:coverage}, the faithfulness score for the unlabeled dialogue is as follows,
\begin{equation}
\label{eq:faith_factcc}
\lambda_{fai}=\Phi(\tilde{V}^b)
\end{equation}
Operation $\Phi$ will be introduced in Sec.~\ref{sec:vec2score}.

\begin{figure}[tbh]
\centering
\includegraphics[width=0.48\textwidth]{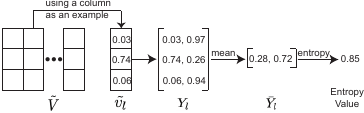}
\caption{
The diagram of the variant-length multi-label BNN. It uses a $\tilde{V}$ column as an example to obtain an entropy value. This example sets $k=3$. The $\lambda_{cov/fai}$ is the sum of the entropy values from all $\tilde{V}$ columns.
}
\label{fig:ml_bnn}
\end{figure}

\subsection{Mean and Bayesian Neural Network}
\label{sec:vec2score}
Once we have a coverage matrix $\tilde{V}^t\in\mathbb{R}^{k \times p}$ or a faithfulness matrix $\tilde{V}^b\in\mathbb{R}^{k \times h}$, we propose three types of operations $\Phi$ to calculate coverage score or faithfulness score, which all measure prediction quality. For simplicity, we let $\tilde{V}$ denote $\tilde{V}^t$ or $\tilde{V}^b$.
\\
\textbf{Mean.} As a straightforward method~\cite{zhang2024don}, we consider the mean value of $\tilde{V}$ to be the required scalar score, that is, coverage or faithfulness score $\lambda_{cov/fai}=mean(\tilde{V})$. 
In this case, $\lambda_{cov/fai}$ represents the average score across all $k$ diverse summaries, measuring coverage or faithfulness.
\\
\textbf{Multi-label BNN.} However, the mean operation only considers the values themselves but ignores the distribution between $[\tilde{v}_{1}, \tilde{v}_{2}, ..., \tilde{v}_{k}]$. BNN has been an effective distribution-based uncertainty estimation method and is suitable for quality assessment~\cite{gawlikowski2023survey}. 
However, BNN 
is originally designed for the fixed-length single-label case, while our case is for variant-length multi-label. Concretely, each element in $\tilde{v}$ can be taken as a label (a noun or a key point) and a good summary should belong to each label, thus our case is multi-label. Besides, each dialogue might have different sizes of nouns ($p$) and key points ($h$), and hence our case has variant lengths. 
Thus, we propose a multi-label BNN to consider the distribution as below, 
\begin{equation}
\begin{split}
\label{eq:multibnn}
\lambda_{cov/fai}=&\sum_{l=1}^{p~(or~h)}\mathbb{H}[Y_l|x, D]
\end{split}
\end{equation}
where we treat the multi-label problem as $p$ or $h$ binary single-label problem. Concretely, we first conduct min-max normalization on $\tilde{V}$. Then, shown as in Figure~\ref{fig:ml_bnn}, we get  $Y_l\in\mathbb{R}^{k\times 2}$ by $\tilde{v_l}\in\mathbb{R}^{k}$ which is a column of the $\tilde{V}$. $Y_l$ is a concatenation of $k$ vector $[\tilde{V}_{i,l}, 1-\tilde{V}_{i,l}]$ for $i$-th predicted summary in $l$-th label. 
Eq.~\ref{eq:multibnn} first calculates the mean $\bar{Y}_l \in \mathbb{R}^{1\times 2}$ of $Y_l\in\mathbb{R}^{k\times 2}$ for $l$-th label, followed by calculating the entropy of $\bar{Y}_l$. Finally, the sum of all $p$ or $h$ labels' entropy values is $\lambda_{cov/fai}$. We use sum rather than mean because our task is variant length and we expect all labels' uncertainty to be accumulated.
More detail is in Sec.~\ref{sec:app_bnn}. 
\\
\textbf{m+BNN.} The mean operation only considers the prediction quality from the view of logits. The BNN method only considers the prediction quality from the view of uncertainty estimation. Therefore, we propose m+BNN to multiply them as below,
\begin{equation}
\label{eq:meanbnn}
\lambda_{cov/fai}= mean(\bar{V}) \times \sum_{l=1}^{p~(or~h)}\mathbb{H}[Y^l|x, D]
\end{equation}

\begin{table*}[]
\caption{Table of data splitting. The numbers in brackets are the sample size. }
\label{tab:datasplit}
\centering
\small
\begin{tabular}{l|cc|cc}
\hline
          & \multicolumn{2}{c|}{Data split setting with small-size labeled data}        & \multicolumn{2}{c}{Data split setting with medium-size labeled data}      \\
          & \multicolumn{1}{c}{Labeled data size} & \multicolumn{1}{c|}{Unlabeled data size} & \multicolumn{1}{c}{Labeled data size} & \multicolumn{1}{c}{Unlabeled data size} \\
          \hline
SAMSUM    & 1\% (147)                        & 50\% (7366)                       & 5\% (736)                        & 50\% (7366)                       \\
DIALOGSUM & 1\% (124)                        & 50\% (6230)                       & 5\% (623)                        & 50\% (6230)                       \\
TODSUM    & 2\% (157)                        & 90\% (7103)                       & 10\% (789)                       & 90\% (7103)   
\\
\hline
\end{tabular}
\end{table*}

\subsection{Fine-Tune on Unlabeled Dialogues Selected by SiCF Scores}
Once we have obtained $\lambda_{sein}$, $\lambda_{cov}$, and $\lambda_{fai}$, we would like to merge them into a SiCF score. However, these three scores are in different scales. Therefore, we use the permutation of pseudolabels based on each of these three scores.

Concretely, given a $\lambda \in \{\lambda_{sein}, \lambda_{cov}, \lambda_{fai}\}$, we sort the pseudolabels in descending order based on $\lambda$. As a result, an order number $\delta$ is positively correlated to the quality of pseudolabel, because a lower $\lambda$ means higher quality. To calculate the SiCF score, we then proceed as follows,
\begin{equation}
\label{eq:sicf}
\lambda_{SiCF}=(\alpha \delta_{SeIn}+\beta \delta_{cov}+\gamma \delta_{fai}) / 3N
\end{equation}
where $\alpha$, $\beta$, and $\gamma$ are hyperparameter. A larger $\lambda_{SiCF}$ means higher quality. As shown in Figure~\ref{fig:global_view}, we select a certain ratio (e.g., 25\%) of unlabeled dialogue by the SiCF scores on the chosen predicted summaries, which is described in Sec.~\ref{sec:overview_ssds}. 
Finally, we fine-tune $G_{0}$ on selected <unlabeled dialogues, pseudolabel> and all labeled pairs to get our target dialogue model $\hat{G}$.

\section{Experiments}

\subsection{Task setting}
Our goal is to select a certain ratio of <unlabeled dialogues, pseudolabels> with high SiCF scores, which are used to fine-tune the baseline model together with all labeled samples. In addition, we are interested in the effectiveness of SiCF score from a view of uncertainty estimation.

\begin{table*}[!htb]
    \caption{Uncertainty estimation results of SiCF score on three datasets in terms of BERTScore-F. The ``0-50'' and ``0-90'' are the mean of BERTScore-F with eliminated ratio range 0\%-50\% and 0\%-90\%. In the medium-size setting, the TODSUM has a mean and standard deviation as shown in Table~\ref{tab:std_mean_todsum_medium_uncertainty_BertF_merge}.} 
    \label{tab:uncertainty_BertF_merge}
    \centering
    \scriptsize
\begin{tabular}{l|cc|cc|cc|cc|cc|cc}
\hline
\multirow{2}{*}{ROUGE-1} & \multicolumn{2}{c|}{SAMSUM(1:50)} & \multicolumn{2}{c|}{DIALOGSUM(1:50)} & \multicolumn{2}{c|}{TODSUM(2:90)} & \multicolumn{2}{c|}{SAMSUM(5:50)} & \multicolumn{2}{c|}{DIALOGSUM(5:50)} & \multicolumn{2}{c}{TODSUM(10:90)}\\
                         & 0-50\%     & 0-90\%       & 0-50\%      & 0-90\%        & 0-50\%     & 0-90\%
                         & 0-50\%     & 0-90\%       & 0-50\%      & 0-90\%        & 0-50\%     & 0-90\%
                         \\ \hline
Random Rank    & 57.92 & 69.13 & 58.03 & 69.21 & 77.66 & 83.65 & 59.39 & 70.21 & 59.46 & 70.24 & 81.05 & 86.11 \\
SiCF(mean)     & 58.90 & 70.28 & 58.85 & 70.08 & 78.78 & 84.80 & 60.45 & 71.44 & 60.42 & 71.29 & 82.05 & 87.17 \\
SiCF(BNN)      & 59.34 & 70.64 & \bf59.57 & \bf70.82 & 78.51 & 84.64 & 60.87 & 71.80 & \bf61.03 & \bf71.93 & 82.08 & 87.27 \\
SiCF(mean+BNN) & 59.38 & 70.69 & 59.25 & 70.46 & 78.85 & 84.91 & 60.80 & 71.74 & 60.78 & 71.68 & 82.23 & 87.38 \\
SiCF(m+BNN-s)           &\bf{59.47}	&\textbf{70.78}	&59.37	&70.61	&\bf{78.95}	&\bf{84.97}  &\bf{60.95}	&\bf{71.89}	&60.74	&71.68	&\bf{82.27}	&\bf{87.43}  \\
\hline
Pseudo Oracle                                     & 61.35 & 72.88 & 61.59 & 72.98 & 81.56 & 87.70 & 62.95 & 74.07 & 62.98 & 74.00 & 84.98 & 90.16
\\
\hline
\end{tabular}
\end{table*}

\subsection{Data}
We conduct experiments on three datasets: (1) SAMSUM~\cite{gliwa2019samsum} is a daily chat domain dataset with 14732 training samples, 818 evaluation samples and 819 testing samples. (2) DIALOGSUM~\cite{chen2021DIALOGSUM} is a dataset in real-life scenarios with training, evaluation, and testing sample sizes of 12460, 500, and 500, respectively. (3) TODSUM~\cite{zhao2021todsum} is a dataset with task-oriented dialogues in seven domains with training, evaluation, and testing sample sizes of 7460, 999, and 999, respectively. We split each dataset into labeled and unlabled portions, and experiment with two settings 
(\textbf{small} and \textbf{medium}-size labeled setting) controlling for the size of the labeled data as shown in Table~\ref{tab:datasplit}.

\begin{table*}[!htb]
    \caption{SSDS results on SAMSUM and TODSUM. 
    The values in the bracket are SSDS improved ratios. ROUGE-L is listed in Tables~\ref{tab:ssds_merged_wiL1} and~\ref{tab:ssds_merged_wiL2}. In the medium-size setting, the TODSUM has a mean and standard deviation as shown in Table~\ref{tab:mean_std_todsum_medium_ssds_merged}. The SSDS results on DIALOGSUM are listed in Table~\ref{tab:ssds_DIALOGSUM_wiL}.
    } 
    \label{tab:ssds_merged}
    \centering
    \small
\begin{tabular}{l|ccc|ccc}
\hline
\multirow{2}{*}{} & \multicolumn{3}{c|}{Small-Size Labeled Data} & \multicolumn{3}{c}{Medium-Size Labeled Data} \\
\cline{2-7}
                   & {ROUGE-1} & {ROUGE-2} & {BERTScore-F}  & {ROUGE-1} & {ROUGE-2} & {BERTScore-F} \\
\hline
\hline
\multicolumn{7}{c}{SAMSUM}
\\
\hline
Initial Fine-Tuned                                                    & 43.90(0\%)  & 18.49(0\%)    & 43.74(0\%)  & 46.81(0\%)  & 20.46(0\%)   & 45.74(0\%)                                                               \\
\hline
Full Unlabeled                                                                & 44.32(41\%)  & 19.07(42\%)    & 43.72(-3\%)  & 47.76(62\%)  & 21.84(91\%)    & \bf{47.02}(\bf{81\%)}                                            \\
Random Rank        & 44.98(105\%)  & 19.32(60\%)    & 44.39(112\%)  & 47.65(55\%)  & 21.13(44\%)    & 46.66(58\%)                                              \\
SiCF (mean)                                                             & \bf{45.85}(\bf{191\%)}  & 19.90(102\%)    & \bf{44.89}(\bf{198\%)}  & 47.90(71\%)  & 21.39(61\%)    & 46.51(48\%)                                              \\
SiCF (BNN)                                                             & 45.20(127\%)  & \bf{19.95}(\bf{105\%)}    & 44.45(122\%)  & 47.77(63\%)  & \bf{22.07}(\bf{106\%)}    & 46.35(38\%)                                                                               \\
SiCF (m+BNN)                                                         & 45.14(121\%)  & 19.31(59\%)  & 44.47(125\%)  & \bf{48.14}(\bf{87\%)}  & 22.05(105\%)    & 46.61(55\%)                                               \\
SiCF (m+BNN-s)    & 45.40(147\%)  & 19.38(64\%)    & 44.34(103\%)  & 47.83(67\%)  & 21.40(62\%)    & 46.55(51\%)                                              \\
\hline
Pseudo Oracle                                                          & 44.92(100\%)  & 19.87(100\%)    & 44.32(100\%)  & 48.33(100\%)  & 21.97(100\%)    & 47.32(100\%) 
\\                                             \hline 
\hline
\multicolumn{7}{c}{TODSUM}
\\
\hline
Initial Fine-Tuned                                                        & 76.60(0\%)  & 59.51(0\%)   & 70.02(0\%)  & 79.75(0\%)  & 64.69(0\%)  & 74.28(0\%)                          \\
\hline
Full Unlabeled                                                                  & 77.02(16\%)  & 59.86(9\%)   & 70.57(13\%)  & 80.00(6\%)  & 64.99(5\%)  & 74.30(0\%)                                 \\
Random Rank                                                            & 76.91(12\%)  & 59.43(-2\%)   & 70.64(15\%)  & 80.57(22\%)  & 65.73(19\%)   & 75.13(20\%)                                                  \\
SiCF (mean)                                                             & \bf{77.94}(\bf{53\%)}  & \bf{61.01}(\bf{40\%)}    & \bf{71.72}(\bf{43\%)}  & 81.08(36\%)  & 66.70(37\%)    & 75.75(34\%)                                              \\
SiCF (BNN)                                                              & 76.64(1\%)  & 59.66(4\%) & 70.53(12\%)   & \bf{82.01}(\bf{61\%)}  & \bf{67.90}(\bf{59\%)}   & \bf{76.74}(\bf{58\%)}                                               \\
SiCF (m+BNN)                                                        & 77.01(16\%)  & 59.92(11\%) & 70.91(22\%)  & 80.93(32\%)  & 66.32(30\%)   & 75.78(35\%)                                                \\
SiCF (m+BNN-s)                                                         & 76.45(-6\%)  & 59.46(-1\%)   & 70.64(15\%)  & 81.22(39\%)  & 67.14(45\%)   & 76.15(44\%)                                                       \\
\hline
Pseudo Oracle                                                           & 79.09(100\%)  & 63.19(100\%)   & 73.96(100\%)  & 83.43(100\%)  & 70.10(100\%)  & 78.48(100\%)                                               
\\
\hline

\end{tabular}
\end{table*}

\subsection{Baselines}

We compare our proposed method to two baselines: (1) \textbf{Random Rank}: a method using the same chosen sampling ratio but using a random rank, (2) \textbf{Full Unlabeled}: a method using all unlabeled dialogues without any selection, which has been verified to be a strong baseline in selective  learning~\cite{he2023zero}. We also calculate an upper bound for ranking termed \textbf{pseudo oracle}, where the ground truth summary is used to score the pseudo summaries according to the BERTScore-F.

\subsection{Metrics}
\textbf{Metric for uncertainty estimation}, force-truth evaluation, evaluates the quality score from a view of uncertainty estimation~\cite{zhang2019mitigating,he2020towards,he2023clur}. Concretely, it simulates the performance improvement of quality scores with human involvement. We measure F-values of BERTScore, ROUGE-1, 2, and L at different elimination ratios. Concretely, for $N$ testing samples and an elimination ratio $r$, the most uncertain predicted summaries in size of $N\times r$ are set as ground truth summaries. The more accurate the uncertainty scores we obtain, the more inaccurate predicted summaries will be replaced by ground truth summaries under the same $r$, resulting in a better summarization metric (e.g., ROUGE-1) score. The dialogue summary metric score at a 0\% eliminated ratio represents the original model's summarization performance. The summary metric scores at 10\%-90\% elimination ratios reflect the uncertainty estimation results. We report the mean of metric performance between 0-50\% and between 0-90\%. 
\\
\textbf{Metrics for SSDS} are BERTScore~\cite{zhang2019bertscore}, ROUGE-1, -2, and -L~\cite{lin2004ROUGE}. Besides, since we have calculated pseudo oracle, we can get \textbf{SSDS improved ratios}, which indicates the improved degree compared to pseudo oracle. A SSDS improved ratio (IR) is defined as below, which is similar to normalized WER in ~\citet{gu2023scaling}, 
\begin{equation}
\label{eq:SSDS-IR}
IR = \frac{MS_{m}-MS_{ini}}{MS_{ora}-MS_{ini}}
\end{equation}
where $MS_{m}$ is a method's metric score, $MS_{ini}$ is the initial finie-tuned's metric score, and $MS_{ora}$ is the pseudo oracle's metric score. A higher improved ratio signifies method superiority, further explained in Sec.~\ref{sec:app_ssds_imp_ra}.

\subsection{Experimental Setting}
We first only use all labeled dialogues to fine-tune the pretrained DialougeLED~\cite{zhong2022dialoglm} model, which is called \textbf{initial fine-tuned}.
For SSDS, we set the selected ratio of unlabeled dialogues as 25\% by default. The full unlabeled method uses all (100\%) unlabeled dialogues. $k=20,8,8$ for SAMSUM, DIALOGSUM, and TODSUM respectively.
For uncertainty estimation, we rank all unlabeled dialogues and replace the most uncertain summaries with their ground truth summaries with a given ratio from [0\%, 10\%, ..., 90\%]. We set the 
$\alpha$, $\beta$, $\gamma$ to 1 by default, where parameter analysis and search (based on ROUGE-1) are presented in Sec.~\ref{sec:paramter_ana_sea}. We abbreviate our searched m+BNN results as \textbf{m+BNN-s}. The repetitive experimental setting is detailed in Sec.~\ref{sec:app_robust_setting}.

\subsection{Experimental Results}

\subsubsection{Uncertainty Estimation Results}

We present our uncertainty estimation results on SiCF scores in Table~\ref{tab:uncertainty_BertF_merge} and~\ref{tab:uncertainty_R1_merge}. In these tables, we only list the ROUGE-1 and BERTScore-F. The details of ROUGE-1, -2, -L, and BERTScore-F are drawn in Figure~\ref{fig:unc_samsum_1p},~\ref{fig:unc_samsum_5p},~\ref{fig:unc_dialog_1p},~\ref{fig:unc_dialog_5p},~\ref{fig:unc_todsum_1p} and~\ref{fig:unc_todsum_5p} in appendix. With these tables, we conclude as below.

\textbf{SiCF score is an effective way to improve uncertainty estimation.} In the two tables, the SiCF scores always outperform the random rank baseline. For example, in Table~\ref{tab:uncertainty_BertF_merge}, all six settings indicate that the SiCF (m+BNN) improves at least 1 point BERTScore-F compared with random rank in both small and medium-size labeled settings. 
This shows that SiCF effectively quantifies the quality of generated summaries by semantic invariance, coverage, and faithfulness, surpassing the straightforward outcomes of the random baseline.

\textbf{SiCF (m+BNN) performs better than SiCF (mean) and SiCF (BNN) in the vast settings.} The difference between the three designs is slight, with less than 1 point divergence. However, SiCF (m+BNN) performs better than SiCF (mean) and SiCF (BNN) in the vast settings, except the ROUGE-1 on TODSUM (2:90) in Table~\ref{tab:uncertainty_R1_merge}. For BERTScore-F in Table~\ref{tab:uncertainty_BertF_merge},  BNN performs better than m+BNN on DIALOGSUM, which verifies the effectiveness of our designed variant-length multi-label BNN. 
Nevertheless, the better performance of SiCF (m+BNN) in the other two datasets still indicates that a m+BNN is better in uncertainty estimation in general.

\subsubsection{SSDS Results}

Table~\ref{tab:ssds_merged},~\ref{tab:ssds_DIALOGSUM_wiL} list SSDS results. We conclude below.

\textbf{SiCF score is effective to improve SSDS.} 
On the two settings of SAMSUM, the three ways of SiCF are generally beneficial for the SSDS compared with random rank. Though the improvement is less than 1 point, our SiCF shows a much better SSDS improved ratio (198\%) compared to random rank (112\%) and full unlabeled in terms of improved ratio in BERTScore-F. 
For the medium-size labeled settings of DIALOGSUM, only SiCF (m+BNN) generally performs better than the random rank but SiCF (mean) and SiCF (BNN) do not. This indicates that the combination of mean and multi-label BNN is effective. 
We do not report SSDS results on DIALOGSUM 1:50, as its generated pseudolabels are too noisy for training, which is detailed in the caption of Table~\ref{tab:ssds_merged_50ratio}.
As for the two settings of TODSUM, 
SiCF (BNN) generally improves at least 2 point in terms of ROUGE-1, 2 and L compared to random rank in the medium setting. These verify the effectiveness of our SiCF score in improving SSDS by selecting unlabeled dialogues via SiCF scores.

\textbf{Our methods surpasses pseudo oracle due to higher sample diversity.} As a surprising finding, in Table~\ref{tab:ssds_merged}, our SiCF (m+BNN) is higher than pseudo oracle in terms of ROUGE-1 (e.g., 45.14 VS. 44.92) and BERTScore-F in both SAMSUM 1:50 and DIALOGSUM 5:50 settings.  The possible reason is that the initial fine-tuned model might be good at predicting a certain distribution of the unlabeled dialogues. As a result, selected unlabeled dialogues in the pseudo oracle may be less diverse than those from SiCF (m+BNN).

\textbf{Using all the unlabeled dialogues is not the best choice because some samples have significant pseudolabel noise.} Compared to the results of full unlabeled, all metrics generally indicate that selecting 25\% high-quality unlabeled dialogues is better. This is because only a part of generated pseudolabels is beneficial to the SSDS learning, which is verified in Table~\ref{tab:ssds_merged_50ratio}.
Thus, it is essential to select high-quality <dialogue, pseudolabel> pairs.

\subsubsection{Ablation Studies}

We list our ablation studies in Tables~\ref{tab:ab_samsum_unc_R1_1p_part},~\ref{tab:ab_samsum_unc_R1_1p}, and~\ref{tab:ab_samsum_unc_bertf_1p} 
for SAMSUM datasets with 1:50 setting. In the tables, ``SiCF'' method uses all three components. ``Only sein'', ``Only cov'', and ``Only fai'' are methods only using respective component. The two table answer the below question.

\textbf{Among the three components in the SiCF score, the coverage score performs better, while a combination of three parts achieves overall improvement.} Concretely, based on the two tables,   using three components together generally improves performance except in SiCF mean cases. Its possible reason is that the pretrained model used in faithfulness might not perform well in SiCF, as the hallucination in text summarization is still a challenging problem. But we believe future research could improve the hallucination detection. 
As a result, SiCF using BNN and m+BNN both have results when using all three components.

\begin{table}[!htb]
    \caption{Ablation study of three components in SAMSUM 1:50 in terms of ROUGE-1. We use $\pm$ to connect the mean and standard deviation among 4 times repetitive experiments with different random seeds. The full table is in Tables~\ref{tab:ab_samsum_unc_R1_1p}.
    } 
    \label{tab:ab_samsum_unc_R1_1p_part}
    \centering
    \small
\begin{tabular}{l|cc}
\hline
\multirow{2}{*}{ROUGE-1}                                    & \multicolumn{2}{|c}{SiCF (m+BNN)}                               \\
              &            \multicolumn{1}{|l}{0-50\% mean} & \multicolumn{1}{l}{0-90\% mean} \\
                         \hline
sein+cov+fai & \bf59.932$\pm$0.015 & \bf71.151$\pm$0.031 \\
only sein     & 59.533$\pm$0.011 & 70.675$\pm$0.010  \\
only cov      & 59.811$\pm$0.010  & 71.022$\pm$0.005 \\
only fai     & 59.104$\pm$0.004 & 70.037$\pm$0.004 
\\
\hline
\end{tabular}
\end{table}

\subsubsection{Parameter Analysis \& Human Eval}
\label{sec:paramter_ana_sea}

We conduct parameter analysis on TODSUM (2:90) with SiCF (m+BNN), where its SSDS result is in Table~\ref{tab:pa_ssds_tod_2_90}, and its uncertainty estimation results are in Table~\ref{tab:pa_unc_todsum_2_90}. Plus, the parameter search results are shown in Table~\ref{tab:coef_hp_search}.
For more details, please refer to Sec.~\ref{sec:parameter_analysis}. Human evaluation is in Sec.~\ref{sec:app_human_eval}.

\section{Conclusion}

To make use of unlabeled data and measure generated summary quality for summarization model training, we benchmark SSDS and uncertainty estimation on dialogue summarization. We propose the SiCF score, which measures semantic invariance, coverage, and faithfulness of pseudolabels (generated summaries) at the text level, word level, and sentence level, respectively. 
Furthermore, we extend BNN-based uncertainty estimation to a variant-length multi-label setting.
Our SiCF score can enhance uncertainty estimation on dialogue summarization and improve SSDS by up to +1-2\% ROUGE and BERTScore-F on SAMSUM (daily-chat domain),  TODSUM (task-oriented dialogues) and DIALOGSUM (real-life scenario). 

\section{Ethical Consideration}
This study pioneers the evaluation of summary quality without relying on ground truth summaries. During our study, we address the challenge of diverse ground truth summaries for dialogue summarization in an innovative way.

Our research employs datasets that are publicly available, ensuring transparency and accessibility. The datasets integral to our work are utilized in adherence to their respective licenses, which is verified in Sec. \ref{sec:license}.

All the datasets utilized in our study are devoid of personal identification details. We advise that any potential extensions of this research into domains containing personal or sensitive information should be conducted with strict adherence to robust ethical guidelines.

\section{Limitations}
This paper introduces the SiCF score as a means of evaluating the quality of generated summaries without using ground truth summaries. However, SiCF has a limitation: when we restrict the occurrence of proper nouns to a maximum of one, it affects other proper nouns that are not speaker names. Therefore, we need to explore alternative approaches for weighting each key detail and key point.

\bibliography{main}

\clearpage
\appendix

\section{Appendix}
\begin{table*}[!htb]
    \caption{SSDS results on TODSUM 2:90 and 10:90 settings with 50\% ratio. From the table, we can see that the pseudo oracle using 25\% unlabeled data performs better than the pseudo oracle using 50\% unlabeled data, and even better than that using full (100\%) unlabeled data. Due to the high level of noise in the pseudolabels of the DIALOGSUM dataset in the small-sized labeled setting, the pseudo oracle compared to the initially fine-tuned model shows only marginal improvement. Consequently, we choose to skip experiments on DIALOGSUM (1:50). } 
    \label{tab:ssds_merged_50ratio}
    \centering
    \scriptsize
\begin{tabular}{l|c|cccc|cccc}
\hline
\multirow{2}{*}{} & \multirow{2}{*}{Select Ratio} & \multicolumn{4}{c|}{1:50 or 2:90} & \multicolumn{4}{c}{5:50 or 10:90} \\
\cline{3-10}
                   &  & {ROUGE-1} & {ROUGE-2} & {ROUGE-L} & {BERTScore-F}  & {ROUGE-1} & {ROUGE-2} & {ROUGE-L} & {BERTScore-F} \\
\hline
\hline
\multicolumn{10}{c}{SAMSUM}
\\
\hline
Initial fine-tuned & N/A                                                        & 43.90                                               & 18.49                                               & 35.02                                               & 0.4374                                                           & { 46.81}                        & { 20.46}                        & { 36.38}                        & {45.74}                                                                         \\
\hline
Full Unlabeled            & 100\%                                                        & 44.32                                               & 19.07                                               & 35.43                                               & 43.72                                                            & { 47.76}                        & { 21.84}                        & \bf{38.20}                        & {47.02}                                              \\
Pseudo Oracle      & 50\%                                                      & \bf{45.58}                                               & 19.79                                               & \bf{35.95}                                               & 44.00                                                          & \bf{48.40}                                               & 21.73                                               & 38.16                                               & \bf{47.57}                                                  \\
Pseudo Oracle      & 25\%                                                     & 44.92                                               & \bf{19.87}                                               & 35.67                                               & \bf{44.32}                                                       & 48.33                                               & \bf{21.97}                                               & 37.93                                               & 47.32
\\                                             \hline 
\hline
\multicolumn{10}{c}{DIALOGSUM}
\\
\hline
Initial Fine-Tuned & N/A                                                       & { 40.30}                        & { 14.53}                        & { 31.19}                        & {43.89}  & { 42.28}                        & { 15.61}                        & { 33.18}                        & { 46.29}                            \\
\hline
Full Unlabeled            & 100\%                                                        & { 39.22}                        & { 14.04}                        & { 30.50}                        & {43.49}  & { 41.61}                        & { 16.04}                        & { 33.25}                        & {46.12}                          \\
Pseudo Oracle      & 50\%                                                      & 40.37                                               & 14.54                                               & 31.44                                               & 44.69      & 42.44                                               & \bf{16.37}                                               & \bf{33.80}                                               & 46.87                                            \\
Pseudo Oracle      & 25\%                                                     & \bf{40.44}                                               & \bf{14.98}                                               & \bf{31.73}                                               & \bf{45.43}   & \bf{42.71}                                               & 16.10                                               & 33.77                                               & \bf{47.05}                                            
\\
\hline
\hline
\multicolumn{10}{c}{TODSUM}
\\
\hline
Initial Fine-Tuned & N/A                                                        & { 76.60}                        & { 59.51}                        & { 67.87}                        & {70.02}  & 79.75                      & 64.69                        & 72.77                        &74.28                         \\
\hline
Full Unlabeled            & 100\%                                                        & { 77.02}                        & { 59.86}                        & { 68.09}                        & {70.57} & 80.00                        & 64.99                        & 73.08                        & 74.30                                \\
Pseudo Oracle      & 50\%                                                      & 78.44                                               & 61.65                                               & 70.11                                               & 72.37  & 82.49                                               & 68.48                                               & 76.40                                               & 77.18                                                \\
Pseudo Oracle      & 25\%                                                     & \bf{79.09}                                               & \bf{63.19}                                               & \bf{71.94}                                               & \bf{73.96}    & \bf{83.43}                                               & \bf{70.10}                                               & \bf{77.43}                                               & \bf{78.48}                                              
\\
\hline

\end{tabular}
\end{table*}

\subsection{More related work}
\label{app:rel_sum_qua}
\noindent\textbf{Semi-supervised text summarization.} There are two general types of text summarization: extractive text summarization \cite{wong2008extractive,liu2019fine}, which extracts the original sentences from the text for summarization, and abstractive text summarization \cite{liu2019text,nallapati2016abstractive}, which directly generates summaries from the given text. For semi-supervised text summarization, \citet{sahu2023enchancing} acknowledge that it is heavily under-explored.

In terms of semi-supervised extractive text summarization, \citet{sahu2023enchancing} use a prompt-based method to implement extractive summarization, but their prompt is based on a subset of extracted sentences rather than the full text. In contrast, our work considers the full text.

In terms of abstractive text summarization, \citet{wang2023t5} employ transfer learning for semi-supervised abstractive text summarization instead of quality assessment. While \citet{gidiotis2023bayesian} assess the quality of the generated summary, they measure the disagreement between each pair of generated summaries for a text via Monte Carlo (MC) dropout \cite{gal2016dropout}. In contrast, our semantic invariance has a lower time complexity due to no comparison among each pair. Additionally, our coverage and faithfulness consider the relation between dialogues and generated summaries, while \citet{gidiotis2023bayesian} do not.

Also related to unlabeled text, \citet{tsvigun2023active} focus on active learning via the similarity between labeled and unlabeled text. In contrast, we focus on semi-supervised learning, and our SiCF score considers the relation between pseudo labels and unlabeled dialogues.

\noindent\textbf{Summary quality.} 
Summary quality measurement can be divided into: with ground truth and without ground truth. Previous quality measurement methods mostly rely on availability of ground truth, such as ROUGE Scores~\cite{lin2004ROUGE} and BERT Scores~\cite{zhang2019bertscore}. Additionally, criteria like coverage and faithfulness were proposed as metrics to evaluate summarization in~\citet{huang2023swing}, which also uses ground truth summary. Faithfulness can also be evaluated~\citet{dreyer2023evaluating} through automated evaluation, as proposed in FactCC~\cite{kryscinski2021evaluating} and 
DOCNLI~\cite{yin2021docnli}. Besides, coherence is proposed in~\citet{zhong2022towards}.
However, these evaluations all require ground truth summaries. In contrast, the SSDS task is inaccessible to ground truth, and the measurement of summary quality without relying on ground truth is underexplored.
Although semantic uncertainty was introduced to assess generation quality from the view of uncertainty~\cite{kuhn2023semantic}, it is time-consuming and overlooks the relationship between the generation and context. Consequently, we propose a more efficient method for measuring semantic invariance. Moreover, we are the first to comprehensively evaluate summary quality focusing on semantic invariance, coverage, and faithfulness, without relying on ground truth summaries.

\subsection{Model}
\subsubsection{Reasons for choosing semantic invariance, coverage, and faithfulness}
\label{sec:app_resons_three}
Inspired by~\citet{huang2023swing}, our justification for choosing semantic invariance, coverage, and faithfulness is based on three key requirements. Firstly, we aim to perform quality analysis within the generated summaries (semantic invariance) and evaluate the quality between the generated summaries and the quality analysis (coverage and faithfulness). Secondly, we aim to conduct quality evaluations at the token level (coverage), sentence level (faithfulness), and text level (semantic invariance). Thirdly, we strive for our quality evaluation to be inaccessible to ground truth summaries, which are often unavailable in real-world scenarios. Our designed three components fulfill these three requirements. 

Additionally, our coverage and faithfulness evaluations do not rely on ground truth summaries, whereas the coverage and faithfulness evaluations in ~\citet{huang2023swing} do.

\subsubsection{Comparison between POS and NER model}
\label{sec:compare_pos_ner}
To extract key information from the dialogues, the intuitive choice is the Named Entity Recognition (NER) model, while we find that most of the extracted entities from 'ner-english-ontonotes' are all speaker names. In contrast, we find that nouns can represent the key information of the dialogues in a good way. 

An example of comparing the use of an NER model and a Part-of-Speech (POS) model to extract key information is shown below, where the \underline{blue font} indicates the annotated key information.

\noindent 1. Dialogue: "\textcolor{blue}{Tom}: Happy \textcolor{blue}{B-day}! Tom: <file gif> \textcolor{blue}{Laura}: oh , thank you , it's so cute <3 <3 <3 Tom: :D"

\noindent 2. Ground truth summaries: \textcolor{blue}{Tom} wishes \textcolor{blue}{Laura} happy \textcolor{blue}{brithday}. 

\noindent 3. Predicted summaries:\textcolor{blue}{Laura} thanked \textcolor{blue}{Tom} for his \textcolor{blue}{birthday}. 

The \underline{NER model extracted results} are shown as below:

\noindent 1. Dialogue: \{Tom: Person, Tom: Person, Laura: Person\}

\noindent 2. Ground truth summaries: \{Tom: Person, Laura: Person\}

\noindent 3. Predicted summaries: \{Tom: Person, Laura: Person\}

Here, we observe that the extracted entities are all person names. Given that the dialogues often contain many speaker names, which cannot effectively differentiate between different dialogues, we believe that using a NER model to extract key information may not be ideal.

The \underline{POS model extracted results} are shown below, where NNP represents Proper noun, singular, and NN represents Noun, singular or mass.

\noindent 1. Dialogue: \{Tom: NNP, B-day: NN, Tom: NNP, file: NN, Laura: NNP, Tom: NN, D: NN\}

\noindent 2. Ground truth summaries: \{Tom: NNP, Laura: NNP, B-day: NN\}

\noindent 3. Predicted summaries:  \{Tom: NNP, Laura: NNP, B-day: NN\}

In this case, we find that the POS results for the dialogue have a large overlap with the key information from a human perspective (the blue fonts in the original dialogue).  Thus, the nouns in the dialogue are a better representation of key information compared to the NER model.

\subsubsection{Multi-label BNN}
\label{sec:app_bnn}

\begin{equation}
\begin{split}
\label{eq:multibnn_detailed}
\lambda_{cov/fai}=&\underbrace{\sum_{l=1}^{p/h}\mathbb{H}[Y_l|x, D]}_{Predictive}=
\underbrace{\sum_{l=1}^{p/h}\mathbb{I}[Y_l|x, D]}_{Epistemic} + \\
&\underbrace{\sum_{l=1}^{p/h}\mathbb{E}[\mathbb{H}[Y_l|x, D]]}_{Aleatoric}
\end{split}
\end{equation}

In BNN theory~\cite{gawlikowski2023survey}, predictive uncertainty usually consists of aleatoric uncertainty and epistemic uncertainty. 

The aleatoric uncertainty is irreducible because it refers to the noise in data generation, such as imperfect sensors. And the epistemic uncertainty refers to a model uncertainty due to limited knowledge, such as having insufficient training data. The predictive BNN is the sum of these two items.

To get the three kinds of uncertainty, we usually firstly calculate the predictive uncertainty as the mean $\bar{Y}_l$ of $Y_l\in\mathbb{R}^{k\times 2}$ for $l$-th label, followed by calculating the entropy of $\bar{Y}_l$. Finally, the sum of all labels' entropy is the predictive uncertainty score in terms of coverage $\lambda_{cov}$ and faithfulness score $\lambda_{fai}$. To obtain aleatoric uncertainty, we calculate the entropy of each $Y_{l}$ at first, before calculating the expectation of all $k$ entropy values; finally, the sum of all expectations of $l$ labels' entropy expectation is the aleatoric uncertainty score. As for epistemic uncertainty, it is usually obtained by using a predictive uncertainty score to subtract its epistemic uncertainty.

\begin{table*}[!htb]
    \caption{SSDS results on SAMSUM 1:50 setting and on TODSUM 2:90 setting. 
    The values in the bracket are SSDS improved ratios. 
    } 
    \label{tab:ssds_merged_wiL1}
    \centering
    \small
\begin{tabular}{l|cccc}
\hline
\multirow{2}{*}{} & \multicolumn{4}{c}{1:50 or 2:90} \\
\cline{2-5}
                   & {ROUGE-1} & {ROUGE-2} & {ROUGE-L} & {BERTScore-F}  \\
\hline
\hline
\multicolumn{5}{c}{SAMSUM}
\\
\hline
Initial Fine-Tuned                                                    & 43.90(0\%)  & 18.49(0\%)  & 35.02(0\%)  & 43.74(0\%)                                                 \\
\hline
Full Unlabeled                                                                & 44.32(41\%)  & 19.07(42\%)  & 35.43(63\%)  & 43.72(-3\%)                                              \\
Random Rank        & 44.98(105\%)  & 19.32(60\%)  & 35.65(96\%)  & 44.39(112\%)                                                \\
SiCF (mean)                                                             & \bf{45.85}(\bf{191\%)}  & 19.90(102\%)  & \bf{35.96}(\bf{144\%)}  & \bf{44.89}(\bf{198\%)}                                                \\
SiCF (BNN)                                                             & 45.20(127\%)  & \bf{19.95}(\bf{105\%)}  & 35.63(93\%)  & 44.45(122\%)                                                                                \\
SiCF (m+BNN)                                                         & 45.14(121\%)  & 19.31(59\%)  & 35.59(87\%)  & 44.47(125\%)                                                 \\
SiCF (m+BNN-s)    & 45.40(147\%)  & 19.38(64\%)  & 35.48(70\%)  & 44.34(103\%)                                            \\
\hline
Pseudo Oracle                                                          & 44.92(100\%)  & 19.87(100\%)  & 35.67(100\%)  & 44.32(100\%)  
\\                                             \hline 
\hline
\multicolumn{5}{c}{TODSUM}
\\
\hline
Initial Fine-Tuned                                                        & 76.60(0\%)  & 59.51(0\%)  & 67.87(0\%)  & 70.02(0\%)                    \\
\hline
Full Unlabeled                                                                  & 77.02(16\%)  & 59.86(9\%)  & 68.09(5\%)  & 70.57(13\%)                   \\
Random Rank                                                            & 76.91(12\%)  & 59.43(-2\%)  & 68.29(10\%)  & 70.64(15\%)                                               \\
SiCF (mean)                                                             & \bf{77.94}(\bf{53\%)}  & \bf{61.01}(\bf{40\%)}  & \bf{69.64}(\bf{43\%)}  & \bf{71.72}(\bf{43\%)}                                    \\
SiCF (BNN)                                                              & 76.64(1\%)  & 59.66(4\%)  & 68.71(20\%)  & 70.53(12\%)                         \\
SiCF (m+BNN)                                                        & 77.01(16\%)  & 59.92(11\%)  & 69.01(28\%)  & 70.91(22\%)                                                \\
SiCF (m+BNN-s)                                                         & 76.45(-6\%)  & 59.46(-1\%)  & 68.77(22\%)  & 70.64(15\%)                                                    \\
\hline
Pseudo Oracle                                                           & 79.09(100\%)  & 63.19(100\%)  & 71.94(100\%)  & 73.96(100\%)                                    
\\
\hline

\end{tabular}
\end{table*}

\begin{table*}[!htb]
    \caption{SSDS results on SAMSUM 5:50 setting and on TODSUM 10:90 setting. 
    The values in the bracket are SSDS improved ratios. 
    } 
    \label{tab:ssds_merged_wiL2}
    \centering
    \small
\begin{tabular}{l|cccc}
\hline
\multirow{2}{*}{} & \multicolumn{4}{c}{5:50 or 10:90} \\
\cline{2-5}
                    & {ROUGE-1} & {ROUGE-2} & {ROUGE-L} & {BERTScore-F} \\
\hline
\hline
\multicolumn{5}{c}{SAMSUM}
\\
\hline
Initial Fine-Tuned                                                      & 46.81(0\%)  & 20.46(0\%)  & 36.38(0\%)  & 45.74(0\%)                                                               \\
\hline
Full Unlabeled                                                                  & 47.76(62\%)  & 21.84(91\%)  & \bf{38.20}(\bf{117\%)}  & \bf{47.02}(\bf{81\%)}                                            \\
Random Rank          & 47.65(55\%)  & 21.13(44\%)  & 37.18(51\%)  & 46.66(58\%)                                              \\
SiCF (mean)                                                               & 47.90(71\%)  & 21.39(61\%)  & 37.57(76\%)  & 46.51(48\%)                                              \\
SiCF (BNN)                                                              & 47.77(63\%)  & \bf{22.07}(\bf{106\%)}  & 37.67(83\%)  & 46.35(38\%)                                                                               \\
SiCF (m+BNN)                                                           & \bf{48.14}(\bf{87\%)}  & 22.05(105\%)  & 37.63(80\%)  & 46.61(55\%)                                               \\
SiCF (m+BNN-s)     & 47.83(67\%)  & 21.40(62\%)  & 37.24(55\%)  & 46.55(51\%)                                              \\
\hline
Pseudo Oracle                                                           & 48.33(100\%)  & 21.97(100\%)  & 37.93(100\%)  & 47.32(100\%) 
\\                                             \hline 
\hline
\multicolumn{5}{c}{TODSUM}
\\
\hline
Initial Fine-Tuned                                                      & 79.75(0\%)  & 64.69(0\%)  & 72.77(0\%)  & 74.28(0\%)                          \\
\hline
Full Unlabeled                                                          & 80.00(6\%)  & 64.99(5\%)  & 73.08(6\%)  & 74.30(0\%)                                 \\
Random Rank                                                             & 80.57(22\%)  & 65.73(19\%)  & 73.76(21\%)  & 75.13(20\%)                                                  \\
SiCF (mean)                                                             & 81.08(36\%)  & 66.70(37\%)  & 74.48(36\%)  & 75.75(34\%)                                              \\
SiCF (BNN)                                                              & \bf{82.01}(\bf{61\%)}  & \bf{67.90}(\bf{59\%)}  & \bf{75.95}(\bf{68\%)}  & \bf{76.74}(\bf{58\%)}                                               \\
SiCF (m+BNN)                                                        & 80.93(32\%)  & 66.32(30\%)  & 74.58(38\%)  & 75.78(35\%)                                               \\
SiCF (m+BNN-s)                                                          & 81.22(39\%)  & 67.14(45\%)  & 75.16(51\%)  & 76.15(44\%)                                                       \\
\hline
Pseudo Oracle                                                           & 83.43(100\%)  & 70.10(100\%)  & 77.43(100\%)  & 78.48(100\%)                                               
\\
\hline

\end{tabular}
\end{table*}

\begin{table*}[!htb]
    \caption{SSDS results on DIALOGSUM 5:50 settings. We do not report SSDS results on DIALOGSUM 1:50, as its generated pseudolabels are too noisy for training and detailed in caption of Tab.~\ref{tab:ssds_merged_50ratio}.
    } 
    \label{tab:ssds_DIALOGSUM_wiL}
    \centering
    \small
\begin{tabular}{l|cccc}
\hline
\multirow{2}{*}{} & \multicolumn{4}{c}{Medium-size Labeled Data} \\
\cline{2-5}
                   & {ROUGE-1} & {ROUGE-2} & {ROUGE-L} & {BERTScore-F}  \\
\hline
\hline
\multicolumn{5}{c}{DIALOGSUM}
\\
\hline
Initial Fine-Tuned                                                       
 
& 42.28(0\%)  & 15.61(0\%)  & 33.18(0\%)  & 46.29(0\%)          \\
\hline
Full Unlabeled                                                                   
& 41.61(-155\%)  & 16.04(87\%)  & 33.25(11\%)  & 46.12(-22\%)           \\
Random Rank                                                             

& 42.32(9\%)  & 16.38(157\%)  & 33.70(88\%)  & 46.07(-28\%)  \\
SiCF (mean)                                                             

& 41.77(-118\%)  & 15.87(53\%)  & 32.79(-66\%)  & 45.48(-106\%)                                               \\
SiCF (BNN)                                                            

& 42.00(-65\%)  & 15.95(69\%)  & 33.47(49\%)  & 46.70(53\%)                                                \\
SiCF (m+BNN)                                                         

& 42.85(132\%)  & 16.86(255\%)  & 34.06(149\%)  & 46.45(21\%)                                               \\
SiCF (m+BNN-s)                                                        

& \bf{43.02}(\bf{172\%)}  & \bf{17.22}(\bf{328\%)}  & \bf{34.32}(\bf{193\%)}  & \bf{47.02}(\bf{96\%)} 
                                       \\
\hline
Pseudo Oracle                                                           

& 42.71(100\%)  & 16.10(100\%)  & 33.77(100\%)  & 47.05(100\%)                                                                         
\\
\hline
\end{tabular}
\end{table*}

\subsubsection{Special Case on Faithfulness}
To benefit the understanding of faithfulness, we omit a special case for faithfulness in Sec.~\ref{sec:fai}, where a dialogue sentence has no nouns. Concretely, when $i$-th dialogue sentence in a dialogue has no nouns, the $i$-th element in $\tilde{v}^{b}\in \mathbb{R}^{h}=min(\tilde{V}^{b})$ will equal 0, because $w_i^b=0$ in Eq.~\ref{eq:point_sim}. We do not expect this to happen for a dialogue sentence without nouns, as a smaller value in $\tilde{v}^{b}$ refers to better faithfulness. Therefore, we add an activation $A_{w^b}$ on $\tilde{v}^{b}$, followed by concatenating $k$ activated results of $A_{w^b}(\tilde{v}^{b})$ to obtain $\tilde{V}^{b}$. 
$A_{w^b}(\tilde{v}^{b})$ is formulated as below, 
\begin{equation}
\label{eq6}
A_{w_i^b}(\tilde{v}_{i}^{b})=\left\{
\begin{aligned}
\tilde{v}_{i}^{b}, & & w^b_i \ge 0 \\
\gamma, & & w^b_i=0
\end{aligned}
\right.
\end{equation}
where it keeps the original $\tilde{v}_{i}^{b}$ if the respective $i$-th sentence in a dialogue has at least one noun, or else gives a large scalar $\gamma$.

\subsection{More Experimental Results}
\subsubsection{More Explanation About SSDS Improved Ratio}
\label{sec:app_ssds_imp_ra}
If a method's improved ratio is greater than 100\%, it indicates superior performance compared to the pseudo oracle. Conversely, if a method's improved ratio is smaller than 0, it suggests worse performance than the initial dialogue summarization that is fine-tuned only on the labeled samples. A higher improved ratio for a method signifies its superiority over another method.

\subsubsection{Uncertainty Estimation Results}
\textbf{For the aleatoric uncertainty is more important in uncertainty estimation than epistemic uncertainty.} In the uncertainty estimation task, based on Table~\ref{tab:alea_epis_unc_sam_1_50_1} and ~\ref{tab:alea_epis_unc_sam_1_50_2}, we see that aleatoric performs better than epistemic in both SiCF (BNN) and SiCF (m+BNN). This indicates that the aleatoric uncertainty (such as noise in the sample collection impacts more than the epistemic uncertainty (such as insufficient training samples). It further shows that pseudolabel noise impacts more compared with the unlabeled sample size in SSDS on the SAMSUM 1:50 setting. Also, in SiCF (BNN), using both aleatoric and epistemic (our default usage) improves the uncertainty estimation. In contrast, in SiCF (m+BNN), only using both aleatoric  (our default usage) improves the uncertainty estimation. The possible reason is that the mean information is more complementary to aleatoric compared with epistemic. However, in our experiments, we still use the predictive uncertainty, a sum of aleatoric and epistemic uncertainty, which is a commonly usage of the two kinds of uncertainty.

\begin{table}[!htb]
\caption{Human evaluation results on the TODSUM on 100 testing samples from the SSDS task in the medium-size setting, assessed by 5 participants.} 
    \label{tab:human_tod_medium}
    \centering
    \small
\begin{tabular}{l|c}
\hline
Method    & \multicolumn{1}{l}{Human preference rate $\uparrow$} \\
\hline
Full Unlabeled & 21.20\%                                   \\
SiCF(mean)      & 23.40\%                                   \\
SiCF(BNN)       & 22.80\%                                   \\
SiCF(m+BNN)     & \textbf{32.60\%}                           \\ \hline
\end{tabular}
\end{table}

\begin{table*}[!htb]
    \caption{Comparison of aleatoric and epistemic of uncertainty estimation results on SAMSUM 1:50 in terms of ROURE-1 and ROUGE-2. } 
    \label{tab:alea_epis_unc_sam_1_50_1}
    \centering
    \small
\begin{tabular}{l|cc|cc}
\hline
                                        & \multicolumn{2}{c|}{ROUGE-1}                                       & \multicolumn{2}{c}{ROUGE-2}                                                                  \\
                            
\multirow{-2}{*}{}                      & \multicolumn{1}{l}{0-50\% mean} & \multicolumn{1}{c|}{0-90\% mean} & \multicolumn{1}{l}{0-50\% mean} & \multicolumn{1}{c}{0-90\% mean}  \\ \hline
SiCF(BNN, alea+epis)      & \bf{59.84}                           & \bf{71.03}                           & \bf{40.34}                           & \bf{56.66}                                                     \\
SiCF(BNN, alea)           & 59.80                           & 70.98                           & 40.30                                                    & 56.58                          \\
SiCF(BNN, epis)           & 59.70                           & 70.85                           & 40.21                           & 56.53                                                    \\
\hline
SiCF(m+BNN, alea+epis) & 59.91                           & 71.10                           & 40.41                           & 56.77                                                     \\
SiCF(m+BNN, alea)      & \bf{59.94}                           & \bf{71.17}                           & \bf{40.41}                           & \bf{56.78}                                                     \\
SiCF(m+BNN, epis)      & 59.83                           & 70.99                           & 40.35                           & 56.67                                                   
\\
\hline
\end{tabular}
\end{table*}

\begin{table*}[!htb]
    \caption{Comparison of aleatoric and epistemic of uncertainty estimation results on SAMSUM 1:50 in terms of ROUGE-L and BERTScore-F. } 
    \label{tab:alea_epis_unc_sam_1_50_2}
    \centering
    \small
\begin{tabular}{l|cc|cc}
\hline
& \multicolumn{2}{c|}{ROUGE-L}                                       & \multicolumn{2}{c}{BERTScore-F}                                   \\
                            
\multirow{-2}{*}{}                       & \multicolumn{1}{l}{0-50\% mean} & \multicolumn{1}{c|}{0-90\% mean} & \multicolumn{1}{l}{0-50\% mean} & \multicolumn{1}{l}{0-90\% mean} \\ \hline
SiCF(BNN, alea+epis)                                 & \bf{52.69}                           & \bf{65.87}                           & 59.34                          & \bf{70.64}                          \\
SiCF(BNN, alea)                                     & 52.68                           & 65.83                           & \bf{59.35}                          & \bf{70.64}                          \\
SiCF(BNN, epis)                                     & 52.42                           & 65.58                           & 59.09                          & 70.40                          \\
\hline
SiCF(m+BNN, alea+epis)                            & 52.62                           & 65.80                           & 59.38                          & 70.69                          \\
SiCF(m+BNN, alea)                                & \bf{52.68}                           & \bf{65.89}                           & \bf{59.45}                          & \bf{70.78}                          \\
SiCF(m+BNN, epis)                            & 52.53                           & 65.69                           & 59.24                          & 70.53                         
\\
\hline
\end{tabular}
\end{table*}

\begin{figure*}[!htbp]
\centering
\includegraphics[width=\textwidth]{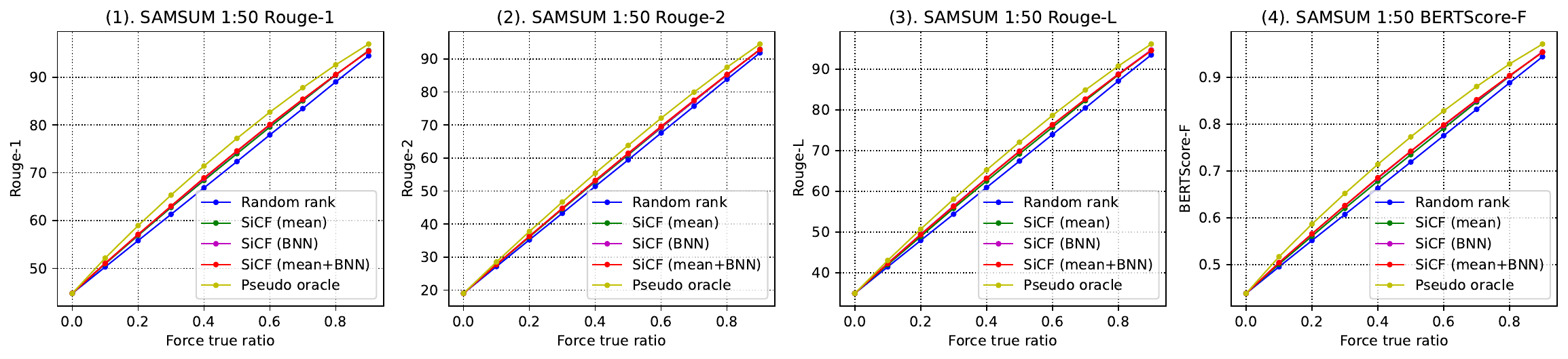}
\caption{Diagram of uncertainty estimation results in force true ratio of 0\%, 10\%, 20\% ..., 90\% on SAMSUM 1:50 setting.}
\label{fig:unc_samsum_1p}
\end{figure*}

\begin{figure*}[!htbp]
\centering
\includegraphics[width=\textwidth]{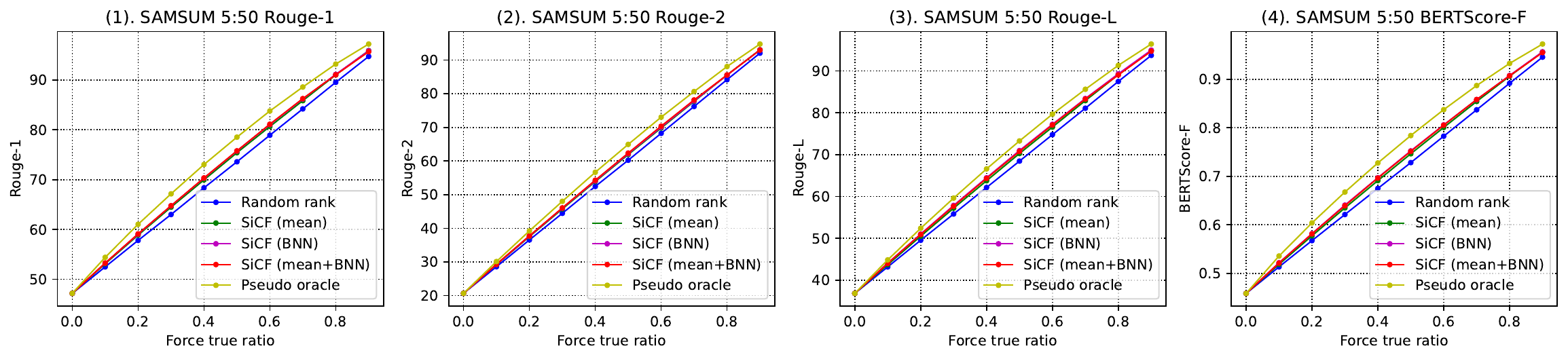}
\caption{Diagram of uncertainty estimation results in force true ratio of 0\%, 10\%, 20\% ..., 90\% on SAMSUM 5:50 setting.}
\label{fig:unc_samsum_5p}
\end{figure*}

\begin{figure*}[!htbp]
\centering
\includegraphics[width=\textwidth]{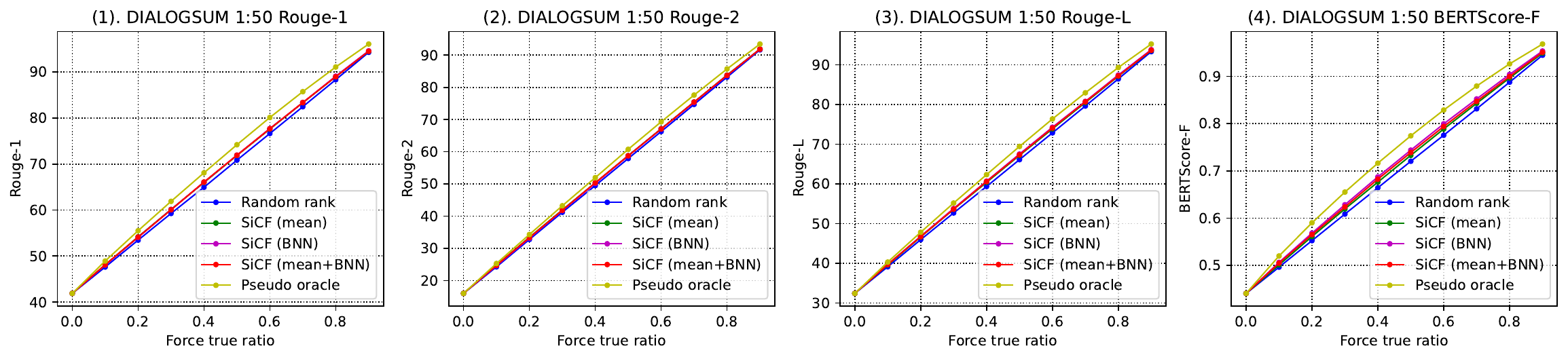}
\caption{Diagram of uncertainty estimation results in force true ratio of 0\%, 10\%, 20\% ..., 90\% on DIALOGSUM 1:50 setting.}
\label{fig:unc_dialog_1p}
\end{figure*}

\begin{figure*}[!htbp]
\centering
\includegraphics[width=\textwidth]{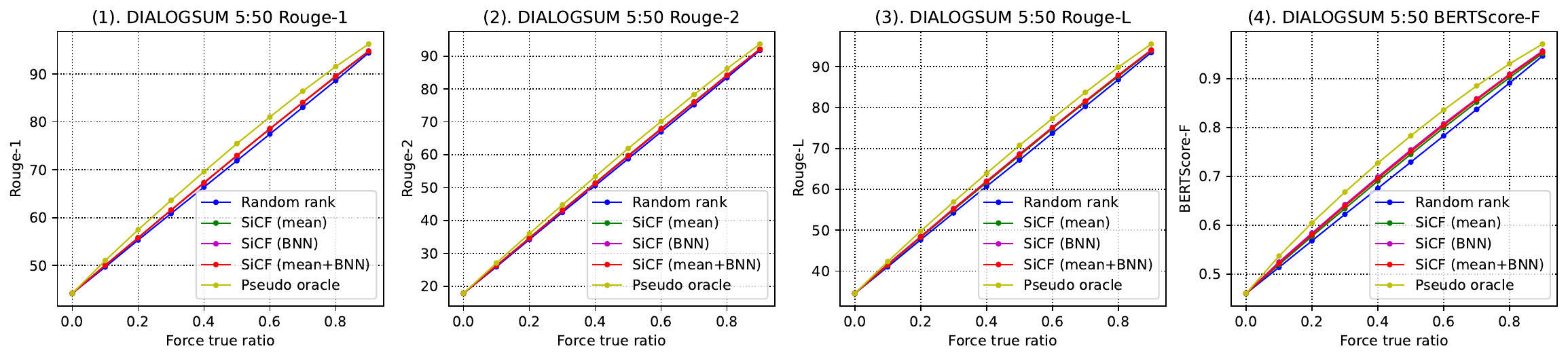}
\caption{Diagram of uncertainty estimation results in force true ratio of 0\%, 10\%, 20\% ..., 90\% on DIALOGSUM 5:50 setting.}
\label{fig:unc_dialog_5p}
\end{figure*}

\begin{figure*}[!htbp]
\centering
\includegraphics[width=\textwidth]{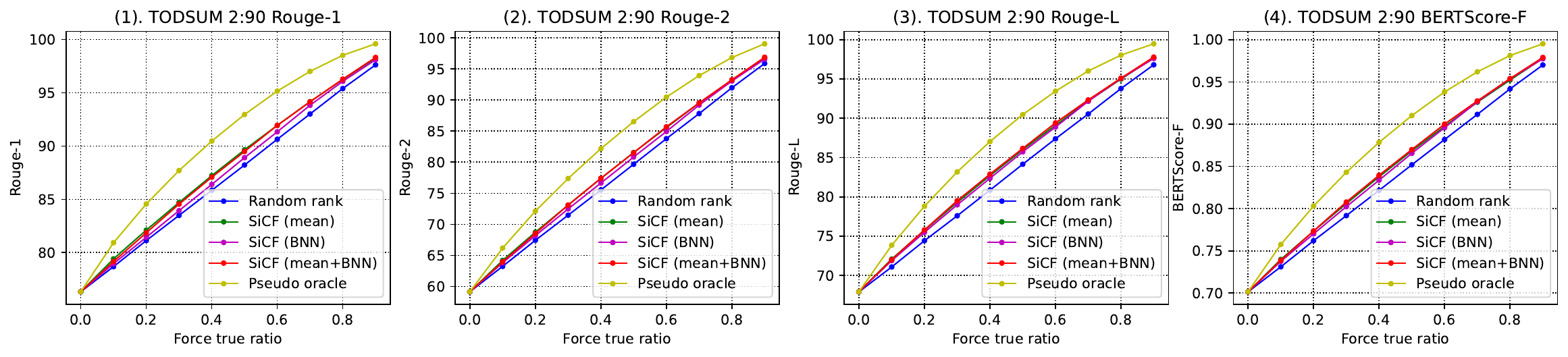}
\caption{Diagram of uncertainty estimation results in force true ratio of 0\%, 10\%, 20\% ..., 90\% on TODSUM 2:90 setting.}
\label{fig:unc_todsum_1p}
\end{figure*}

\begin{figure*}[!htbp]
\centering
\includegraphics[width=\textwidth]{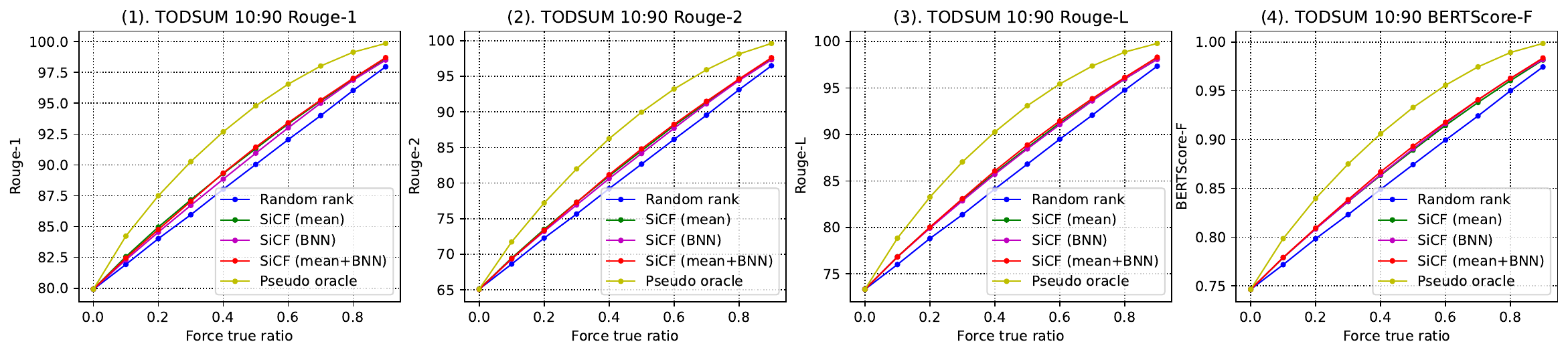}
\caption{Diagram of uncertainty estimation results in force true ratio of 0\%, 10\%, 20\% ..., 90\% on TODSUM 10:90 setting.}
\label{fig:unc_todsum_5p}
\end{figure*}

Besides the listed mean values of 0-50\% and 0-90\% in the Table~\ref{tab:uncertainty_R1_merge} and ~\ref{tab:uncertainty_BertF_merge}, we also draw the concrete metric values in different force true ratios in Figure~\ref{fig:unc_samsum_1p},~\ref{fig:unc_samsum_5p},~\ref{fig:unc_dialog_1p},~\ref{fig:unc_dialog_5p},~\ref{fig:unc_todsum_1p},~\ref{fig:unc_todsum_5p}.

\begin{figure*}[!tbh]
\centering
\includegraphics[width=0.95\textwidth]{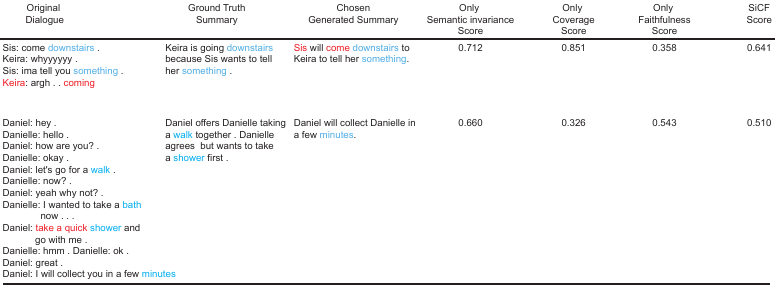}
\caption{
The quality analysis of our SiCF score as well as its ablation study. 
Regarding the semantic invariance score, the $k$ diverse generated summaries are not displayed for brevity. In the coverage score, key details (nouns) are highlighted in blue. For assessing faithfulness, conflicts (first row) and missing information (second row) are indicated in red. The analysis of the figure is in Sec.~\ref{sec:quality_ana_ssds}. Please zoom in for better visualization.
}
\label{fig:mean_bnn_case}
\end{figure*}

\subsubsection{SSDS Results}
\textbf{BNN, and m+BNN performs better than mean in SSDS.} Based on Table~\ref{tab:compare_mean_bnn}, we found among the 12 comparisons, two groups show mean performs better, five groups show that BNN performs better, and five groups show that m+BNN performs better. This demonstrates the positive effect of BNN and a combination of the mean and BNN on SSDS.

\begin{table*}[!htb]
    \caption{Comparison of mean (m), BNN (B), and m+BNN (m+B) on SSDS results based on Table~\ref{tab:ssds_merged}.  ``\textgreater{}'' means better.} 
    \label{tab:compare_mean_bnn}
    \centering
    \small
\begin{tabular}{l|cc|cc|cc}
\hline
                         & \multicolumn{2}{c|}{SAMSUM}                                        & \multicolumn{2}{c|}{DIALOGSUM}                                     & \multicolumn{2}{c}{TODSUM}                                        \\
                         & 1:50                            & 5:50                            & 1:50                            & 5:50                            & 2:90                            & 10:90                           \\
                         \hline
\multicolumn{1}{c|}{50\%} & B\textgreater{}m+B\textgreater{}m & m+B\textgreater{}B\textgreater{}m & B\textgreater{}m+B\textgreater{}m & m+B\textgreater{}B\textgreater{}m & m+B\textgreater{}m\textgreater{}B & B\textgreater{}m+B\textgreater{}m \\
\multicolumn{1}{c|}{25\%} & m\textgreater{}B\textgreater{}m+B & B\textgreater{}m+B\textgreater{}m & m+B\textgreater{}B\textgreater{}m & m+B\textgreater{}B\textgreater{}m & m\textgreater{}m+B\textgreater{}B & B\textgreater{}m+B\textgreater{}m
\\
\hline
\end{tabular}
\end{table*}

\begin{table}[!htb]
    \caption{Mean and standard deviation of uncertainty estimation results of SiCF score on TODSUM datasets in terms of ROUGE-1 on the medium-size labeled data. The results are reported based on four repetitive experiments via different random seeds. The ``0-50'' and ``0-90'' are the mean of ROUGE-1 with eliminated ratio range 0\%-50\% and 0\%-90\%. It is respective to the Table~\ref{tab:uncertainty_R1_merge}.} 
    \label{tab:std_mean_todsum_medium_uncertainty_rouge1_merge}
    \centering
    \small
\begin{tabular}{l|cc}
\hline
\multirow{2}{*}{ROUGE-1} & \multicolumn{2}{c}{TODSUM (10:90)}\\
                         & 0-50\%     & 0-90\%
                         \\ \hline
Random Rank         & 84.963$\pm$0.027 & 88.980$\pm$0.042  \\
SiCF(mean)          & 85.822$\pm$0.028 & 89.823$\pm$0.056 \\
SiCF(BNN)           & 85.509$\pm$0.024 & 89.629$\pm$0.024 \\
SiCF(m+BNN)         & 85.774$\pm$0.033 & 89.887$\pm$0.031 \\
SiCF(m+BNN-s) & \bf85.836$\pm$0.004 & \bf89.949$\pm$0.002 \\
\hline
Pseudo Oracle       & 88.235$\pm$0.003 & 92.298$\pm$0.001
\\
\hline
\end{tabular}
\end{table}

\begin{table}[!htb]
    \caption{Mean and standard deviation of uncertainty estimation results of SiCF score on TODSUM datasets in terms of BERTScore-F on the medium-size labeled data. The results are reported based on four repetitive experiments via different random seeds. The ``0-50'' and ``0-90'' are the mean of BERTScore-F with eliminated ratio range 0\%-50\% and 0\%-90\%. It is respective to the Table~\ref{tab:uncertainty_BertF_merge}.} 
    \label{tab:std_mean_todsum_medium_uncertainty_BertF_merge}
    \centering
    \small
\begin{tabular}{l|cc}
\hline
\multirow{2}{*}{BERTScore-F} & \multicolumn{2}{c}{TODSUM (10:90)}\\
                         & 0-50\%     & 0-90\%
                         \\ \hline
Random Rank         & 81.02$\pm$0.017  & 86.086$\pm$0.032 \\
SiCF(mean)          & 81.935$\pm$0.067 & 87.016$\pm$0.089 \\
SiCF(BNN)           & 82.035$\pm$0.026 & 87.221$\pm$0.028 \\
SiCF(m+BNN)         & 82.198$\pm$0.019 & 87.355$\pm$0.014 \\
SiCF(m+BNN-s) & \bf82.273$\pm$0.002 & \bf87.433$\pm$0.002 \\
\hline
Pseudo Oracle       & 84.977$\pm$0.002 & 90.163$\pm$0.002
\\
\hline
\end{tabular}
\end{table}

\begin{table*}[!htb]
    \caption{Mean and standard deviation of SSDS results on TODSUM medium-size labeled data. The results are reported based on four repetitive experiments via different random seeds. It is respective to Table~\ref{tab:ssds_merged}.
    } 
    \label{tab:mean_std_todsum_medium_ssds_merged}
    \centering
    \small
\begin{tabular}{l|ccc}
\hline
\multirow{2}{*}{} & \multicolumn{3}{c}{Medium-Size Labeled Data} \\
\cline{2-4}
                     & {ROUGE-1} & {ROUGE-2} & {BERTScore-F} \\
\hline
\hline
\multicolumn{4}{c}{TODSUM}
\\
\hline
Full Unlabeled      & 80.19$\pm$0.33 & 65.34$\pm$0.63 & 74.57$\pm$0.41 \\
Random Rank         & 80.66$\pm$0.14 & 66.06$\pm$0.39 & 75.19$\pm$0.2  \\
SiCF(mean)          & 81.49$\pm$0.26 & 67.13$\pm$0.33 & 76.06$\pm$0.22 \\
SiCF(BNN)           & \bf82.28$\pm$0.18 & \bf68.40$\pm$0.32  & \bf77.04$\pm$0.19 \\
SiCF(m+BNN)         & 81.72$\pm$0.47 & 67.47$\pm$0.67 & 76.50$\pm$0.44  \\
SiCF(m+BNN-s) & 81.29$\pm$0.11 & 67.01$\pm$0.17 & 76.05$\pm$0.11 \\
\hline
Pseudo Oracle       & 83.75$\pm$0.25 & 70.52$\pm$0.32 & 78.72$\pm$0.19
\\
\hline

\end{tabular}
\end{table*}

\begin{table*}
    \caption{Uncertainty estimation results of SiCF score on three datasets in terms of ROUGE-1. The ``0-50'' and ``0-90'' are the mean of ROUGE-1 scores with eliminated ratio range 0\%-50\% and 0\%-90\%. In the medium-size setting, the TODSUM has a mean and standard deviation as shown in Table~\ref{tab:std_mean_todsum_medium_uncertainty_rouge1_merge}.} 
    \label{tab:uncertainty_R1_merge}
    \centering
    \scriptsize
\begin{tabular}{l|cc|cc|cc|cc|cc|cc}
\hline
\multirow{2}{*}{ROUGE-1} & \multicolumn{2}{c|}{SAMSUM(1:50)} & \multicolumn{2}{c|}{DIALOGSUM(1:50)} & \multicolumn{2}{c|}{TODSUM (2:90)} & \multicolumn{2}{c|}{SAMSUM(5:50)} & \multicolumn{2}{c|}{DIALOGSUM(5:50)} & \multicolumn{2}{c}{TODSUM(10:90)}\\
                         & 0-50\%     & 0-90\%       & 0-50\%      & 0-90\%        & 0-50\%     & 0-90\%
                         & 0-50\%     & 0-90\%       & 0-50\%      & 0-90\%        & 0-50\%     & 0-90\%
                         \\ \hline
Random Rank              & 58.55          & 69.62            & 56.34           & 67.97              & 82.28          & 87.04  & 60.40                            & 70.97                            & 58.05                            & 69.18                            & 84.98                            & 88.99                            \\
SiCF(mean)               & 59.61          & 70.83            & 57.02           & 68.65              & 83.24          & 88.00  & 61.50                            & 72.24                            & 58.63                            & 69.85                            & 85.87   & 89.92          \\
SiCF(BNN)                & 59.84          & 71.03            & 57.01           & 68.68     & 82.67          & 87.54  & 61.71                            & 72.44                            & 58.61                            & 69.86                   & 85.55                            & 89.67                           \\
SiCF(m+BNN)           & 59.91          & 71.10   & 57.03           & 68.67              & 83.08          & 87.92 & 61.74                            & 72.45                   & 58.64                            & 69.88                            & 85.83                            & 89.94    \\
SiCF(m+BNN-s)           &\bf{59.98}	&\bf{71.20}	&\bf{57.06}	&\bf{68.75}	&\bf{83.19}	&\bf{87.99}  &\bf{61.86}	&\bf{72.61}	&\bf{58.70}	&\bf{69.96}	&\bf{85.84}	&\bf{89.95} \\
\hline
Pseudo Oracle            & 61.63          & 72.98            & 58.41           & 70.35              & 85.49          & 90.32  & 63.56                            & 74.41                            & 60.22                            & 71.66                            & 88.23                            & 92.30          \\
\hline
\end{tabular}
\end{table*}

\textbf{Epistemic uncertainty benefits SSDS performance than aleatoric uncertainty.} From Table~\ref{tab:alea_epis_ssds_sam_1_50}, we see that using both aleatoric and epistemic uncertainty generally leads to better SSDS results. For example, 
Considering Tables~\ref{tab:alea_epis_unc_sam_1_50_1} and ~\ref{tab:alea_epis_unc_sam_1_50_2}, we conclude that though aleatoric uncertainty benefits in improving uncertainty estimation results, using both aleatoric and epistemic uncertainty benefits in improving SSDS results.

\begin{table*}[!htb]
    \caption{Comparison of aleatoric and epistemic of SSDS results on SAMSUM 1:50} 
    \label{tab:alea_epis_ssds_sam_1_50}
    \centering
    \small
\begin{tabular}{l|crrrr}
\hline
  & Select Ratio & \multicolumn{1}{c}{ROUGE-1} & \multicolumn{1}{c}{ROUGE-2} & \multicolumn{1}{c}{ROUGE-L} & \multicolumn{1}{c}{BERTScore-F} \\ \hline
SiCF(BNN, alea+epis)      & 0.25                                 & \bf{45.20}                                               & \bf{19.95}                                               & 35.63                                               & \bf{44.45}                                                  \\
SiCF(BNN, alea)           & 0.25                                 & 44.49                                               & 18.93                                               & 34.77                                               & 43.51                                                  \\
SiCF(BNN, epis)           & 0.25                                 & 45.35                                               & 19.69                                               & \bf{35.77}                                               & 44.39                                                  \\
\hline
SiCF(m+BNN, alea+epis) & 0.25                                 & \bf{45.14}                                               & 19.31                                               & \bf{35.59}                                               & \bf{44.47}                                                  \\
SiCF(m+BNN, alea)      & 0.25                                 & 45.06                                               & 19.02                                               & 35.19                                               & 44.16                                                  \\
SiCF(m+BNN, epis)      & 0.25                                 & 45.07                                               & \bf{19.35}                                               & 35.49                                               & 44.19         
\\
\hline
\end{tabular}
\end{table*}

\subsubsection{Robust Experimental Settings}
\label{sec:app_robust_setting}
To assess the robustness of our experiments, we conducted our experiments four times, in addition to the original single run. Each of these four runs used different random seeds, but they all shared the same initial fine-tuned model and used the same set of beam search sampling to generate summaries. This approach allows us to evaluate the consistency of our SiCF scores in a semi-supervised setting, where the initial fine-tuned model is not our focus and should be consistent for fairness. Furthermore, these repeated experiments employed the same set of beam search sampling generated summaries for a fair assessment of robustness. For example, in the SAMSUM 1:50 setting, all four experiments were provided with the same set of 20 generated summaries for each unlabeled dialogue.
The results of these robustness experiments are presented in Tables~\ref{tab:ab_samsum_unc_R1_1p_part},~\ref{tab:ab_samsum_unc_R1_1p},
\ref{tab:std_mean_todsum_medium_uncertainty_rouge1_merge},~\ref{tab:std_mean_todsum_medium_uncertainty_BertF_merge}, and~\ref{tab:mean_std_todsum_medium_ssds_merged}.
Obtaining Table~\ref{tab:mean_std_todsum_medium_ssds_merged} with an additional 3 rounds of experiments for all 7 methods requires approximately 252 hours (10.5 days) on a server equipped with 4 V100 GPUs.

\subsubsection{Parameter Analysis \& Search}
\label{sec:parameter_analysis}
We conduct a parameter analysis on TODSUM 2:90 with SiCF (m+BNN), where its SSDS result comparison on 25\% ration is shown in Table~\ref{tab:pa_ssds_tod_2_90}, and its uncertainty estimation results are presented in Table~\ref{tab:pa_unc_todsum_2_90}. 

For this section, we can conclude as below.

\textbf{The Impact of Parameters on Uncertainty Estimation and SSDS.} Based on Table~\ref{tab:pa_unc_todsum_2_90}, we observe that enlarging the coefficient of semantic invariance can improve uncertainty estimation results by 0-50\% on TODSUM 2:90. However, the 0-90\% mean performs better when the coefficients of the three components are balanced.

Additionally, according to Table~\ref{tab:pa_ssds_tod_2_90}, the best SSDS performance is achieved when the coefficients of the three components are balanced. This suggests that each component in SiCF scores contributes to the quality measurement of pseudolabels.

\textbf{The best-searched coefficient on the uncertainty estimation.} We listed the best-searched coefficient in Table~\ref{tab:coef_hp_search} by searching the coefficient leading to the best ROUGE-1 on the uncertainty estimation task. The search range is [0, 0.25, 0.5, 0.75, 1] for each of the three component coefficients. Based on the table, there is no 0 coefficient appears. This also indicates that each component benefits the uncertainty estimations in all six settings.

\begin{table*}[!htb]
    \caption{The coefficient of hyperparameter search in terms of ROUGE-1. We use ``sein'' to represent semantic invariance, apply ``cov'' to represent coverage and utilize ``fai'' to denote faithfulness.} 
    \label{tab:coef_hp_search}
    \centering
    \small
\begin{tabular}{l|cc|cc|cc}
\hline
\multicolumn{1}{l|}{} & \multicolumn{2}{c|}{SAMSUM} & \multicolumn{2}{c|}{DIALOGSUM} & \multicolumn{2}{c}{TODSUM} \\
\multicolumn{1}{l|}{} & 1:50         & 5:50        & 1:50          & 5:50          & 2:90        & 10:90        \\
\hline
SeIn                 & 0.5          & 0.5         & 1             & 1             & 0.75        & 0.75         \\
Cov                  & 0.75         & 1           & 1             & 0.5           & 0.25        & 0.5          \\
Fai                  & 0.25         & 0.25        & 0.25          & 0.25          & 0.5         & 0.5    
\\
\hline
\end{tabular}
\end{table*}

\subsubsection{Quality Analysis of SSDS Results}
\label{sec:quality_ana_ssds}
Figure~\ref{fig:mean_bnn_case} presents the quality analysis of SSDS results. Specifically, we observe that the chosen generated summaries in the first row have higher coverage, including terms like ``downstairs'' and ``something.'' However, they differ from the original dialogue in terms of who comes downstairs, resulting in a high coverage score but a low faithfulness score.

In contrast, the generated summaries in the second row lack terms like ``walk,'' ``bath,'' and ``shower,'' leading to a low coverage score. Although they do not have a conflict similar to the first row, they miss the shower-related content, resulting in relatively low faithfulness scores.

\subsubsection{Human Evaluation Results}
\label{sec:app_human_eval}
We had 5 participants for the human evaluation test. We presented each person with the same 100 testing samples, including the original dialogues, ground truth dialogue summaries, and four randomly-ordered generated dialogue summaries from four methods (Full Unlabeled, SiCF (mean), SiCF (BNN), SiCF (mean+BNN)). Subsequently, each person was asked the question ``Among the four generated dialogue summaries, which is the best when compared to the original dialogues and ground truth dialogue summaries?'' 

From Table~\ref{tab:human_tod_medium}, we can see that SiCF (mean + BNN) performs the best from the human perspective with a 32.60\% human preference rate. The human preference rate is the ratio between the total respective selected samples and the total 500 times selection.

\subsubsection{More Experimental Settings}
Our experiments run on 4 V100 GPUs, with 12 hours on SAMSUM for the full training.

\subsection{License Analysis}
\label{sec:license}

The SAMSUM dataset is licensed under CC BY-NC-ND 4.0. The DIALOGSUM dataset is licensed under the MIT License. As for TODSUM, it is publicly released without a specified license. Therefore, our research usage of these datasets complies with their respective licenses.

\begin{table*}[!htb]
    \caption{Ablation study of three components in SAMSUM 1:50 in terms of ROUGE-1. We use ``sein'' to represent semantic invariance, apply ``cov'' to represent coverage and utilize ``fai'' to denote faithfulness. We use $\pm$ to connect the mean and standard deviation among 4 times repetitive experiments with different random seeds. A standard deviation of 0.000 means that differences occur in the decimal places further to the right.  } 
    \label{tab:ab_samsum_unc_R1_1p}
    \centering
    \small
\begin{tabular}{l|cc|cc|cc}
\hline
\multirow{2}{*}{ROUGE-1} & \multicolumn{2}{c|}{SiCF (Mean)}                                   & \multicolumn{2}{c|}{SiCF (BNN)}                                    & \multicolumn{2}{c}{SiCF (m+BNN)}                               \\
                         & \multicolumn{1}{l}{0-50\% mean} & \multicolumn{1}{l|}{0-90\% mean} & \multicolumn{1}{l}{0-50\% mean} & \multicolumn{1}{l|}{0-90\% mean} & \multicolumn{1}{l}{0-50\% mean} & \multicolumn{1}{l}{0-90\% mean} \\
                         \hline
sein+cov+fai & 59.941$\pm$0.191 & 71.139$\pm$0.177 & \bf59.812$\pm$0.016 & \bf70.996$\pm$0.021 & \bf59.932$\pm$0.015 & \bf71.151$\pm$0.031 \\
only sein    & 59.533$\pm$0.011 & 70.675$\pm$0.010  & 59.533$\pm$0.011 & 70.675$\pm$0.010  & 59.533$\pm$0.011 & 70.675$\pm$0.010  \\
only cov     & \bf59.964$\pm$0.006 & \bf71.183$\pm$0.005 & 59.676$\pm$0.005 & 70.812$\pm$0.005 & 59.811$\pm$0.010  & 71.022$\pm$0.005 \\
only fai     & 58.657$\pm$0.026 & 69.794$\pm$0.037 & 59.553$\pm$0.006 & 70.671$\pm$0.005 & 59.104$\pm$0.004 & 70.037$\pm$0.004 
\\
\hline
\end{tabular}
\end{table*}

\begin{table*}[!htb]
    \caption{Ablation study of three components in SAMSUM 1:50 in terms of BERTScore F. The organization is similar to Table~\ref{tab:ab_samsum_unc_R1_1p}.} 
    \label{tab:ab_samsum_unc_bertf_1p}
    \centering
    \small
\begin{tabular}{l|cc|cc|cc}
\hline
\multirow{2}{*}{BERTScore-F} & \multicolumn{2}{c|}{SiCF (Mean)}                                   & \multicolumn{2}{c|}{SiCF (BNN)}                                    & \multicolumn{2}{c}{SiCF (m+BNN)}                               \\
                         & \multicolumn{1}{l}{0-50\% mean} & \multicolumn{1}{l|}{0-90\% mean} & \multicolumn{1}{l}{0-50\% mean} & \multicolumn{1}{l|}{0-90\% mean} & \multicolumn{1}{l}{0-50\% mean} & \multicolumn{1}{l}{0-90\% mean} \\
                         \hline
sein+cov+fai                 & 58.90                          & 70.28                          & \bf{59.34}                          & \bf{70.64}                          & \bf{59.38}                          & \bf{70.69}                          \\
only sein                    & 58.83                          & 70.14                          & 58.83                          & 70.14                          & 58.83                          & 70.14                          \\
only cov                     & \bf{59.21}                          & \bf{70.56}                          & 59.26                          & 70.52                 & 5932                          & 70.63                          \\
only fai                     & 58.01                          & 69.32                 & 59.06                          & 70.31                          & 58.61                          & 69.69     
\\
\hline
\end{tabular}
\end{table*}

\begin{table*}[!htb]
    \caption{Parameter analysis of SSDS results on TODSUM 2:90, where the SiCF scores are all calculated by m+BNN. The values in the brackets are $\alpha$, $\beta$, and $\gamma$, respectively.} 
    \label{tab:pa_ssds_tod_2_90}
    \centering
    \small
\begin{tabular}{l|ccccc}
\hline
                   & \multicolumn{1}{l}{Select Ratio} & \multicolumn{1}{l}{ROUGE-1} & \multicolumn{1}{l}{ROUGE-2} & \multicolumn{1}{l}{ROUGE-L} & \multicolumn{1}{l}{BERTScore-F} \\
                   \hline
SiCF(1, 1, 1)            & 0.25                                 & \bf{77.18}                           & \bf{60.33}                           & \bf{69.35}                           & \bf{71.31}                              \\
SiCF(10, 1, 1)           & 0.25                                 & 76.71                           & 60.07                           & 68.73                           & 71.11                              \\
SiCF(0.1, 1, 1)          & 0.25                                 & 76.67                           & 59.80                           & 68.77                           & 70.80                              \\
SiCF(1, 10, 1)           & 0.25                                 & 76.42                           & 59.42                           & 68.42                           & 70.44                              \\
SiCF(1, 0.1, 1)          & 0.25                                 & 76.96                           & 59.89                           & 68.48                           & 70.81                              \\
SiCF(1, 1, 10)           & 0.25                                 & 76.11                           & 58.41                           & 67.23                           & 69.43                              \\
SiCF(1, 1, 0.1)          & 0.25                                 & 76.24                           & 59.27                           & 67.95                           & 70.41           
\\
\hline
\end{tabular}
\end{table*}

\begin{table*}[!htb]
    \caption{Parameter analysis of uncertainty estimation results on TODSUM 2:90. The values in the brackets are $\alpha$, $\beta$, and $\gamma$, respectively.} 
    \label{tab:pa_unc_todsum_2_90}
    \centering
    \scriptsize
\begin{tabular}{l|cc|cc|cc|cc}
\hline
                                        & \multicolumn{2}{c|}{ROUGE-1}                                       & \multicolumn{2}{c|}{ROUGE-2}                                       & \multicolumn{2}{c|}{ROUGE-L}                                       & \multicolumn{2}{c}{BERTScore-F}                                   \\
                            
\multirow{-2}{*}{}                      & \multicolumn{1}{l}{0-50\% mean} & \multicolumn{1}{c|}{0-90\% mean} & \multicolumn{1}{l}{0-50\% mean} & \multicolumn{1}{c|}{0-90\% mean} & \multicolumn{1}{l}{0-50\% mean} & \multicolumn{1}{c|}{0-90\% mean} & \multicolumn{1}{l}{0-50\% mean} & \multicolumn{1}{l}{0-90\% mean} \\ \hline
SiCF(1, 1, 1)   & 83.08                           & \bf{87.92}                           & 70.61                           & \bf{78.90}                           & 77.38                           & \bf{83.89}                           & 78.85                          & \bf{84.91}                          \\
SiCF(10, 1, 1)  & \bf{83.16}                           & 87.89                           & \bf{70.73}                           & 78.87                           & \bf{77.50}                           & 83.85                           & \bf{78.98}                          & \bf{84.91}                          \\
SiCF(0.1, 1, 1) & 82.85                           & 87.62                           & 70.19                           & 78.45                           & 76.90                           & 83.38                           & 78.42                          & 84.45                          \\
SiCF(1, 10, 1)  & 82.70                           & 87.46                           & 70.14                           & 78.33                           & 77.00                           & 83.43                           & 78.48                          & 84.47                          \\
SiCF(1, 0.1, 1) & 83.14                           & \bf{87.92}                           & 70.60                           & 78.83                           & 77.20                           & 83.66                           & 78.73                          & 84.75                          \\
SiCF(1, 1, 10)  & 82.80                           & 87.50                           & 69.97                           & 78.14                           & 76.45                           & 82.86                           & 78.07                          & 84.03                          \\
SiCF(1, 1, 0.1) & 82.90                           & 87.77                           & 70.44                           & 78.76                           & 77.31                           & 83.84                           & 78.77                          & 84.87      
\\
\hline
\end{tabular}
\end{table*}

\end{document}